\documentclass[lettersize,journal]{IEEEtran}
\usepackage{amsmath,amsfonts}
\usepackage{algorithm}
\usepackage{array}
\usepackage{textcomp}
\usepackage{stfloats}
\usepackage{url}
\usepackage{verbatim}
\usepackage{cite}
\usepackage{colortbl}
\usepackage{xcolor}
\usepackage{pifont}
\usepackage{epsfig}
\usepackage{graphicx}
\usepackage{amssymb}
\usepackage{amsthm}
\usepackage{booktabs}
\usepackage{algpseudocode} 
\usepackage{arydshln}
\usepackage{subfigure}                           
\usepackage{footnote}
\usepackage{threeparttable}
\usepackage{hyperref}
\usepackage{silence}
\usepackage[cmintegrals]{newtxmath}
\newcommand{\ie}{{\emph{i.e.}}}

\newcommand{\eg}{{\emph{e.g.}}}

\newcommand{\etal}{{\emph{et al.}}}
\def\sArt{state-of-the-art}

\newcommand{\RemoveAlgoNumber}{\renewcommand{\fnum@algocf}{\AlCapSty{\AlCapFnt\algorithmcfname}}}
\newcommand{\RevertAlgoNumber}{\algocf@resetfnum}
\newcommand{\gap}{{\textrm{GAP}}}
\newcommand{\figref}[1]{Fig.~\ref{#1}}
\newcommand{\tabref}[1]{Table~\ref{#1}}

\newcommand{\secref}[1]{\ref{#1}}

\newtheorem{theorem}{Theorem}

\newtheorem{property}{Property}

\newcommand{\revise}[1]{\textcolor{black}{#1}}
\newcommand{\up}[1]{\textcolor{black}{#1}}
\newcommand{\down}[1]{\textcolor{black}{#1}}

\usepackage{url,multirow,morefloats,floatflt,cancel,tfrupee}
\begin{document}

\title{BackMix: Regularizing Open Set Recognition by Removing Underlying Fore-Background Priors}

\author{Yu Wang,~\IEEEmembership{Member,~IEEE,}
        Junxian Mu, Hongzhi Huang, Qilong Wang,~\IEEEmembership{Member,~IEEE,} \\Pengfei Zhu, 
        Qinghua Hu,~\IEEEmembership{Senior Member,~IEEE}
\thanks{Yu Wang, Junxian Mu, Hongzhi Huang, Qilong Wang, Pengfei Zhu, and Qinghua Hu are with the College of Intelligence and Computing, Tianjin University, Tianjin 300350, China, and also with the Haihe Laboratory of Information Technology Application Innovation (Haihe Lab of ITAI), Tianjin, China.}
\thanks{Corresponding authors: Pengfei Zhu (zhupengfei@tju.edu.cn) and Qinghua Hu (huqinghua@tju.edu.cn)}

}

\maketitle

\begin{abstract}
    Open set recognition (OSR) requires models to classify known samples while detecting unknown samples for real-world applications. Existing studies show impressive progress using unknown samples from auxiliary datasets to regularize OSR models, but they have proved to be sensitive to selecting such known outliers. 
    In this paper, we discuss the aforementioned problem from a new perspective: Can we regularize OSR models without elaborately selecting auxiliary known outliers?
    We first empirically and theoretically explore the role of foregrounds and backgrounds in open set recognition and disclose that: 1) backgrounds that correlate with foregrounds would mislead the model and cause failures when encounters `partially' known images; 2) Backgrounds unrelated to foregrounds can serve as auxiliary known outliers and provide regularization via global average pooling. Based on the above insights, we propose a new method, Background Mix (BackMix), that mixes the foreground of an image with different backgrounds to remove the underlying fore-background priors. Specifically, BackMix first estimates the foreground with class activation maps (CAMs), then randomly replaces image patches with backgrounds from other images to obtain mixed images for training. With backgrounds de-correlated from foregrounds, the open set recognition performance is significantly improved. The proposed method is quite simple to implement, requires no extra operation for inferences, and can be seamlessly integrated into almost all of the existing frameworks. The code is released on https://github.com/Vanixxz/BackMix.
\end{abstract}

\begin{IEEEkeywords}
Classification, open set recognition, unknown detection, fore-background priors, spurious correlation.
\end{IEEEkeywords}

\section{Introduction}
\IEEEPARstart{C}{onventional} artificial intelligence models primarily tackle visual tasks within closed-set situations, where classes remain consistent throughout both training and testing phases~\cite{zhou2022open}. However, in real-world applications, unknown classes may arise during test, challenging the reliability of the closed-set assumption and leading to an open set scenario~\cite{angiulli2010outlier,basich2022competence,lu2014concept, HANHEIDE2017119}. Open set recognition (OSR) is a task that aims to handle such challenging scenarios, in which the models are required to recognize unknown images while accurately classifying known classes. 

Existing OSR methods can be mainly categorized into three groups: discriminative methods that design open set oriented classification strategies~\cite{bendale2016towards,liang2018enhancing,perera2020generative,liu2020energy,zhou2021learning,xu2023contra}, generative methods that generate input distribution or pseudo-unknown samples to explicitly reserve a certain space for unknown classes~\cite{neal2018open,lee2018a,oza2019c2ae,chen2020learning,chen2021adversarial,yang2022cpl,huang2023cssr}, and auxiliary data-based methods that utilize additional samples from manually selected datasets as available outliers~\cite{hendrycks2018deep,dhamija2018reducing,Perera2019dtl,kong2021opengan}. The significant difference between the first two methods and auxiliary data-based methods lies in whether using available outliers to regularize the model's performance on out-of-distribution data. Many studies show that proper unknown samples can enable models to recognize unknowns better and adapt to open environments with less complexity~\cite{cen2023uosr,bai2024idlike}. 

Although auxiliary data-based methods have impressive performance in unknown detection, we find that they are quite sensitive to the selection of out-of-distribution data on different tasks (See Section~\secref{sec:3.2.2}). Such a problem raises a straightforward question: \textbf{Can we regularize open set recognition models without elaborately selecting auxiliary known outliers?}

In pursuit of this goal, we look into the mechanisms of image recognition. In natural images, foregrounds typically encompass the distinctive regions correlated with categories, while cognitive studies suggest that humans utilize backgrounds as contextual cues during object recognition~\cite{de2018expectations,oliva2007role,ranganath2012two}. For example, cars are typically found on roads but rarely in water, while fish predominantly inhabit the water and are seldom observed on roads. Likewise, image classifiers have been shown to effectively utilize and derive benefit from such underlying priors in object identification~\cite{Barbu2019ObjectNetAL, Sagawa2019DistributionallyRN, Shetty_2019_CVPR, xiao2021noise}. Even more, they can achieve notable performance when foregrounds are masked out~\cite{xiao2021noise}.

To rely on these priors, one has to assume that the classes learned during training will only appear in matching backgrounds during test.
However, in OSR, partial distributions of images may change due to the appearance of unknown classes, in which there are two typical cases: \textbf{1) varying foreground}, \ie, unknown classes arise in known backgrounds. In this case, the model is likely to predict an unknown sample to a known class according to its known background; \textbf{2) varying background}, \ie, known classes arise in rare backgrounds. In this case, the model may mistakenly recognize the foreground due to the unseen information brought by the background~\cite{elephant}. When encountering the above two cases of varying foregrounds or backgrounds, the model may fail to identify test samples due to the `partially' known information. 
Unfortunately, the impact of fore-background priors has been neglected in existing OSR methods, which poses significant challenges to them in tackling the aforementioned cases.

\textbf{In this paper, we delve deeply into the role of fore-background priors in OSR and properly use such information to serve as an effective regularization of unknown classes without auxiliary data.} Firstly, we vary the background distribution during both the training and test phases and empirically find that: 1) The model trained on raw images fails to generalize when known objects appear in unseen backgrounds due to the disruption of the learned fore-background correlations; 2) The model becomes more robust to the presence of unknown backgrounds when the foreground class is de-correlated from the seen backgrounds during training.
Subsequently, we provide a theoretical analysis to explain such findings. We show that backgrounds can serve as known outliers and provide extra regularization via global average pooling (GAP), which is equivalent to using manually selected outliers more flexibly and robustly. 

Based on the above insights, we propose a new method, Background Mix (BackMix), which mixes the foreground of an image with different backgrounds to remove the underlying priors. To avoid precise segmentation annotations and additional costs, we use class activation maps (CAMs)~\cite{Zhou_2016_CVPR} to estimate the foreground region roughly. Then, two random images serve as the target image (TI) and the background image (BI) for mixing, respectively. The processed input is obtained by replacing random patches of TI with background patches of BI. Extensive experiments demonstrate that BackMix enhances both closed-set and open set performance under various evaluation metrics when applied to existing OSR methods or compared to data augmentation techniques. The proposed method is quite simple to implement, does not require additional operation to make inferences, and can be seamlessly integrated into almost all existing OSR frameworks. 

In general, our work has the following contributions:

\begin{enumerate}
    \item We thoroughly discuss the role of fore-background priors and demonstrate that the fore-background priors can mislead the model in OSR. This issue can be resolved by releasing the correlation between foreground and background during training.
    \item We provide insights into the regularization effect of class-unrelated backgrounds, which can enhance open set performance by serving as outliers. Moreover, the internal regularization mechanism is as effective as well-designed auxiliary data-based methods.
    \item We propose BackMix that involves rough foreground estimation using CAMs and mixing up backgrounds from different images to release the inherent correlation.
    \item BackMix is simple to implement and can be seamlessly integrated into other methods. Experimental results show that BackMix significantly improves conventional and \sArt~OSR methods by up to 23.6\% on the AUROC, even enhancing the plain baseline to outperform advanced methods. 
\end{enumerate}

The remainder of this paper is organized as follows. Section \secref{sec:related_work} briefly reviews studies related to this work. Section \secref{sec:pre_method} deeply explores and analyzes the role of backgrounds in OSR. Based on the analysis, we propose and elaborate on a new method in Section \secref{sec:method}. Experimental implementation, metrics and results are provided in Section \secref{sec:exp}. Finally, conclusions and future work are drawn in Section \secref{sec:conclusion}.

\section{\revise{Related Work}}
\label{sec:related_work}
\revise{In this section, we review the literature that relates to our work, mainly including open set recognition methods, data augmentation methods, and researches on spurious correlations.}

\subsection{Open Set Recognition}
Open set recognition aims to detect unknown classes while maintaining accuracy in classifying known classes. In this regard, a similar task out-of-distribution (OOD) detection also attempts to address such a problem. We review related methods and divide them into the following three groups.

 \revise{\textbf{Discriminative methods.} Bendale and Boult~\cite{bendale2016towards} addressed the limitations of SoftMax in OSR by introducing OpenMax, which calibrates classification scores using extreme value theory. \revise{Liang~\etal~\cite{liang2018enhancing} proposed to combine temperature scaling and input preprocessing for improving detection performance without retraining the model.} Perera~\etal~\cite{perera2020generative} developed GDFR, enhancing feature quality with a self-supervised auxiliary task. \revise{Liu~\etal~\cite{liu2020energy} demonstrated that energy scores are more effective than SoftMax scores in distinguishing unknown samples and can be flexibly used as a score function.} Zhou~\etal~\cite{zhou2021learning} employed class and data placeholders to reserve space for unknown classes and adjust overconfident predictions. Xu~\etal~\cite{xu2023contra} used supervised contrastive learning to boost the model's ability to extract robust representations. These methods enhance discriminative power by reinforcing feature learning or implementing tailored classification strategies for open scenarios. However, without specific adaptations for open space, their performance remains limited.}

 \revise{\textbf{Generative methods.} To constrain the boundary between known and unknown, Neal~\etal~\cite{neal2018open} generates images that are close to known classes in latent space. \revise{Lee~\etal~\cite{lee2018a} used a generative classifier and adopted the sore function based on the Mahalanobis distance.} Oza and Patel~\cite{oza2019c2ae} utilized an auto-encoder (AE) as the classifier, identifying unknown samples via reconstruction error. Chen~\etal~\cite{chen2020learning} proposed RPL, a distance-based method using reciprocal points, later refined to ARPL~\cite{chen2021adversarial} with adversarial constraints to limit known class space. Yang~\etal~\cite{yang2022cpl} embedded prototypes of known classes in feature space and replaced SoftMax with a prototype model to exclude unknowns. To represent known classes without devouring, Huang~\etal~\cite{huang2023cssr} developed plugged class-specific AEs at the top of the backbone to generate manifolds for known classes. These methods use generated samples or distributions to model classes and enlarge the discrepancy between known and unknown samples, while the generative modules introduce computational cost and instable performance.}

 \revise{\textbf{Auxiliary data-based methods.} Hendrycks~\etal~\cite{hendrycks2018deep} introduced outlier exposure (OE), training models to assign uniform probabilities to outliers from auxiliary datasets. Dhamija~\etal~\cite{dhamija2018reducing} reduced the intensity of global features in outlier images. Perera and Patel~\cite{Perera2019dtl} developed global negative filters using a large dataset to decrease activation for unknown samples. Recognizing the limitations of available auxiliary data and potential overfitting, Kong and Ramanan~\cite{kong2021opengan} proposed training with both real outliers and generated samples. Cen~\etal~\cite{cen2023uosr} examined the effectiveness of OE, suggesting the inclusion of unknown samples in training for few-shot unified OSR tasks. While auxiliary datasets aid in modeling open space with reduced complexity, they introduce biases tied to outlier distributions, making model performance highly sensitive to the choice of auxiliary data.}

Despite significant progress in OSR, current methods primarily focus on modeling known classes or unknown space, overlooking the impact of image backgrounds. We argue that joint modeling of foregrounds and backgrounds may hinder performance. This insight motivates our new approach and offers valuable perspectives for future OSR studies.
\subsection{Data Augmentation}
\label{sec:2.2}
To mitigate overfitting and enhance the generalization of the model, data augmentation techniques have been extensively employed in various tasks. Current approaches can be broadly categorized into masking-based and mixing-based methods.

\textbf{Masking-based methods.} DeVries and Taylor~\cite{devries2017improved} introduced Cutout to remove random image regions for occlusion-invariant training, while Singh and Lee~\cite{singh2017hide} used random patch hiding to encourage learning from whole objects. To avoid excessive masking, Chen~\etal~\cite{chen2020gridmask} proposed GridMask, which applies grid-pattern masking. These techniques mask image sections while preserving primary objects, reducing the original correlations in training images.

\textbf{Mixing-based methods.} 
Zhang~\etal~\cite{zhang2017mixup} introduced Mixup, combining inputs and labels as linear mixtures of two images. Yun~\etal~\cite{yun2019cutmix} developed Cutmix, which swaps random regions within a batch to create new inputs. Zhou~\etal~\cite{zhou2021mixstyle} mixed images across source domains to generate new styles, enhancing training diversity. We notice these operations can help mix image backgrounds, and this should lead to a positive effect on open set recognition. However, experiments in Section \secref{sec:5.2} show that the improvement in closed-set performance is accompanied by degradation in open set performance—likely due to unintended label mixing during processing, which interferes with predictions in OSR settings.

Data augmentation techniques are generally designed with a closed-set assumption, and their effectiveness in OSR remains largely unexplored. Furthermore, many are heuristic, offering limited reliability across diverse and challenging open scenarios. In this work, we provide both empirical and theoretical insights to establish a feasible and robust approach for OSR.

\subsection{\revise{Spurious Correlations}}
\revise{Spurious correlations occur when models depend on irrelevant or secondary features instead of class-related ones, risking inaccurate predictions if not properly addressed~\cite{ye2024spurious}, we categorize existing methods into two classes:}

\revise{\textbf{Representations enhancement methods.}} 
\revise{Srivastava~\etal~\cite{srivastava2020robustness} used human annotations to capture unmeasured confounders and mitigate distribution shifts. Creager~\etal~\cite{creager2021environment} enabled the model to learn invariant features by incorporating environment inference tasks. Yao~\etal~\cite{yao2022improving} proposed to improve out-of-distribution robustness by augmenting the data with a mixup-based method to learn invariant predictors. These methods aim to enhance the model's ability to capture essential representations by optimizing data or features.}

\revise{\textbf{Debiasing optimization methods.} Du~\etal~\cite{du2023less} proposed to down weight examples with high feature-label bias, reducing the model’s reliance on such shortcuts. Liu~\etal~\cite{liu2023avoiding} reduced spurious correlations by correcting logits to balance group accuracy and minimize bias. Asgari~\etal~\cite{asgari2022masktune} masked learned dominant features, encouraging the model to explore and rely on alternative, unbiased features. 
These methods use specific debiasing optimization objectives to eliminate spurious correlations, making the model more robust.}

\revise{Recently, Ming~\etal~\cite{ming2022impact} analyzed the impact of spurious correlations in OOD detection tasks, the proposed BackMix takes a simple and effective way to address this issue in OSR tasks directly. By mixing diverse backgrounds with foregrounds, BackMix effectively mitigates misleading fore-background priors, improving both closed-set classification accuracy and unknown detection performance without requiring auxiliary data or model component.}

\begin{figure}
  \begin{center}
    \includegraphics[width=1.\linewidth]{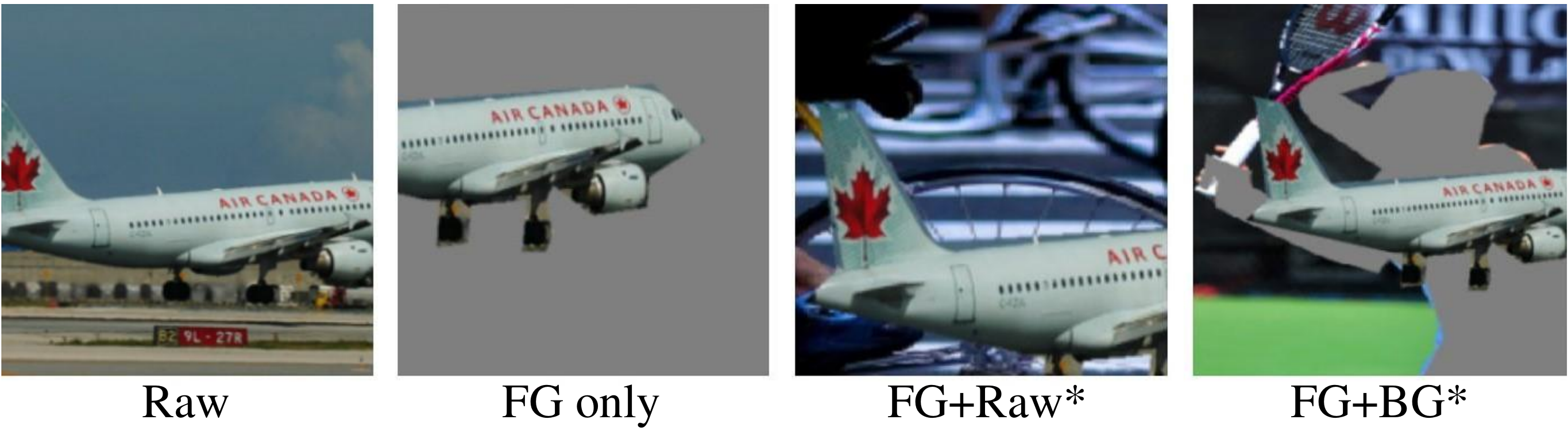}
    \caption{Illustration of operations applied on the generated dataset. Raw, FG, and BG represent the original image, the foreground of the image, and the background of the image, respectively. The star (*) on Raw and BG denotes that another source image is randomly sampled from the dataset.}
    \label{fig:cocotrans}
    \end{center}
\end{figure}

\begin{table*}
    \centering
  \setlength\tabcolsep{10pt}
    \caption{\revise{Performance under various training and test settings. The gain or loss values are calculated compared to the setting I which uses raw images for training. The images used during the test phase are represented as `known test data / unknown test data'.}}
    \begin{tabular}{clllllllll}
    \toprule
         \multirow{2}{*}{Setting} & \multirow{2}{*}{Training Data}
         & \multicolumn{2}{c}{Raw / Raw} & \multicolumn{2}{c}{FG+BG* / Raw} & \multicolumn{2}{c}{FG only / FG only} & \multicolumn{2}{c}{Raw / Raw (iNaturalist)} \\ \cmidrule(r){3-4} \cmidrule(r){5-6} \cmidrule(r){7-8} \cmidrule(r){9-10}
         & & Accuracy & AUROC& Accuracy & AUROC& Accuracy & AUROC& Accuracy & AUROC\\ \midrule
         I & Raw & 80.1 & 67.2 & 69.3 & 59.0 & 72.5 & 64.7 & 80.1 & 74.9\\
         II & FG only & {${17.9}_{\down{62.2\downarrow}}$} & {${52.7}_{\down{14.5\downarrow}}$} & {${30.2}_{\down{39.1\downarrow}}$} & {${53.0}_{\down{6.0\downarrow}}$} & {$92.4_{\up{19.8\uparrow}}$} & {${82.0}_{\up{17.3\uparrow}}$} &{${17.9}_{\down{62.2\downarrow}}$} & {${50.9}_{\down{24.0\downarrow}}$}\\
         III & FG+Raw* & {${73.6}_{\down{6.5\downarrow}}$} & {${68.3}_{\up{1.1\uparrow}}$} & {${85.8}_{\up{16.5\uparrow}}$} & {${83.3}_{\up{24.3\uparrow}}$} & {${89.3}_{\up{16.7\uparrow}}$} & {${75.4}_{\up{10.7\uparrow}}$} & {${73.6}_{\down{6.5\downarrow}}$} & {${85.4}_{\up{10.5\uparrow}}$} \\
         IV & FG+BG* & {${76.4}_{\down{3.7\downarrow}}$} & {${69.5}_{\up{2.3\uparrow}}$} & {${87.3}_{\up{18.0\uparrow}}$} & {${83.6}_{\up{24.6\uparrow}}$} & {${91.5}_{\up{18.9\uparrow}}$} & {${81.0}_{\up{16.3\uparrow}}$} & {${76.4}_{\down{3.7\downarrow}}$} & {${85.6}_{\up{10.7\uparrow}}$} \\
    \bottomrule
    \end{tabular}
    \label{tab:coco}
\end{table*}

\section{Exploring Fore-Background Priors in Open Set Recognition}
\label{sec:pre_method}
In this section, we first explore and discuss the effect of fore-background priors in open set recognition. Based on the analysis, we then provide insights into how class-unrelated backgrounds regularize open set classifiers.

\subsection{Fore-Background Priors in Training Set Misleads Open Set Classifiers}
\label{sec:3.1}
In OSR settings, the model may be faced with partially known test samples. According to the maximum entropy principle~\cite{maximum_entropy_principle}, it is clear that models are supposed to not rely on background information. Training with raw images that have class-related backgrounds, \ie,~some objects appearing only in certain backgrounds, inadvertently introduces prior correlation to the model. These priors may be helpful under the closed-set assumption, where the test sample distribution matches the training ones. However, foreground-background correlations become a hindrance in OSR, potentially misleading classifiers.

To verify the above ideas, we first conducted an experiment to observe the performance trend with varied backgrounds during training and the test.
Following Deng~\etal~\cite{Deng_2021_CVPR}, we used samples from the COCO dataset~\cite{cocods}, which contains pixel-level foreground annotations. We chose 12 classes as the known and 6 classes as the unknown during open set evaluation. With segmentation labels, we generated three variants as follows (See~\figref{fig:cocotrans}). \textit{FG only}: the original background is erased for the image; \textit{FG+Raw*}: the original background is erased and replaced with another random raw image in the dataset; \textit{FG+BG*}: the original background is erased and replaced with a random background of another image.

To observe the impact of fore-background priors, we trained models on the original dataset and the above three variants, respectively. During test, four different settings were considered to evaluate the performance of each model comprehensively:

\begin{enumerate}
    \item Using raw images for known data and unknown data;
    \item Using raw images for unknown data while replacing backgrounds of known data with an unknown image randomly to break the foreground-background correlations;
    \item Using images that have been removed backgrounds for both known and unknown data;
    \item Using raw images for known data while unknown data are from an out-of-distribution dataset iNaturalist~\cite{Horn2018TheIS}, which has little semantic overlap with known classes.
\end{enumerate}

We used the classification accuracy and the Area Under Receiver Operating Characteristic curve (AUROC) to evaluate closed-set and open set performance, respectively. The maximum SoftMax probability~\cite{hendrycks2016a} served as the score function to reject unknown samples. 
Results in~\tabref{tab:coco} show that: 

 \revise{\textbf{Training priors have a negative impact once the correlation breaks down.} As the closed-set performance of learning Raw(Setting I) drops 11\% and 8\% when using the FG+BG* and FG only images as test known samples, the model becomes uncertain about classification without backgrounds and gets worse when given unrelated backgrounds.}

 \revise{\textbf{Simply removing backgrounds is not a practical solution.} As learning FG only \revise{(Setting II)} never considers the existence of backgrounds, its performance seriously degrades when faced with images that have a background. Results suggest that replacing the background of images with pure grey may not be a robust solution for releasing correlations in practice.}

 \revise{\textbf{Releasing fore-background correlations enhances open set performance.}
Learning \emph{FG+Raw*} \revise{(Setting III)} and \emph{FG+BG*} \revise{(Setting IV)} consistently enhance open set performance, which indicates the correlations between foregrounds and backgrounds are hindrances to OSR. Prior experiments showed that using constant backgrounds (pure grey) prevents misleading classifiers but compromises robustness in practice. Therefore, using class-unrelated backgrounds is a feasible way to release the correlations.}

\textbf{Avoiding multiple objects appearing in the foreground improves closed-set performance.} Comparing learning \emph{FG+Raw*} and \emph{FG+BG*}, \emph{FG+BG*} shows better accuracy as it avoids having multiple objects appear in a single image. Therefore, if segmentation is challenging, using raw images to refill the backgrounds exchanges open set performance improvement with a slight closed-set performance drop.

\textbf{Connections to existing opinions.} Previous study on image background shows that models with better classification performance often rely less on backgrounds~\cite{xiao2021noise}. Meanwhile, Vaze~\etal~\cite{vaze2022openset} suggested that a good closed-set classifier inherently enhances open set performance. Our main finding bridges both conclusions, suggesting that good classifiers emphasize foregrounds and are more robust against unknowns.

\subsection{Exploiting Class-Unrelated Backgrounds for Open Set Classifier Regularization}
\label{sec:3.2}
In this section, we first outline two desirable properties that a reliable classifier should have for OSR. Then, we verify that regularizing classifiers with class-unrelated backgrounds shares similarities with OE but with fewer limitations.

\subsubsection{The main theory}
Based on the analysis of foreground-background correlations in OSR, we suppose that models should focus on the foreground objects rather than being misled by backgrounds. To accurately and robustly identify unknowns, with $\mathbf{z}_f$ and $\mathbf{z}_b$ representing the features of foreground and background, a reliable OSR model $\mathcal{W}$ is expected to possess the following two desired properties:

\begin{property}
\label{property1}
The information of foreground and background in any image $\mathbf{x}$ are independent of each other to model $\mathcal{W}$. That is, random variables $\mathbf{z}_f$ and $\mathbf{z}_b$ are independent.
\end{property}

\begin{property}
\label{property2}
For a certain model $\mathcal{W}$, image backgrounds are independent of its prediction on image categories. That is, background feature $\mathbf{z}_b$ and image category $y$ are independent random variables.
\end{property}

Possessing Property~\ref{property1} is mostly beneficial for classification, except for certain special cases, \eg, a person may wear less on the beach. As we have discussed in the previous section, it is difficult for most models to possess the Property~\ref{property2}, but this is quite significant in the open scenarios. 

In modern DNNs, global average pooling (GAP) is widely used~\cite{networkinnetwork}, which is designed to replace intensive fully-connected layers, thereby minimizing overfitting. GAP processes the feature map $\mathcal{Z}\in \mathbb{R}^{H \times W \times C}$ and reduces its spatial dimensions by averaging values to generate a global representation $\mathbf{z}_g=\gap(\mathcal{Z})=\frac{1}{|\mathcal{Z}|}\sum_{\mathbf{z}_{ij}\in\mathcal{Z}}\mathbf{z}_{ij}$. We subsequently explain how DNNs depress background features and regularize networks robust to outliers. Firstly, the global representation $\mathbf{z}_g$ can be decomposed into a linear combination of foreground global features $\mathbf{z}_f$ and background global features $\mathbf{z}_b$:
\begin{align}
    \mathbf{z}_g&=\frac{1}{HW}\sum_{(i,j)\in P}\mathbf{z}_{ij}\notag \\&= \frac{|P_f|}{HW}\times\frac{1}{|P_f|}\sum_{(i,j)\in P_f}\mathbf{z}_{ij}+ \frac{|P_b|}{HW}\times\frac{1}{|P_b|}\sum_{(i,j)\in P_f}\mathbf{z}_{ij} \notag \\
    &=\lambda \times \gap(\mathcal{Z}_f) + (1-\lambda) \times \gap(\mathcal{Z}_b) \notag \\&= \lambda \times \mathbf{z}_f + (1-\lambda) \times \mathbf{z}_b, \label{eq:decompose_gap}
\end{align}
where $P_f$ and $P_b$ are the pixel sets of the foreground or background, respectively, $\lambda$ is the proportion of foreground pixels, and $\mathcal{Z}_*$ is a subset of $\mathcal{Z}$ with elements picked by $P_*$, \ie,~$\mathcal{Z}_*=\{\mathbf{z}_{ij} \mid (i,j)\in P_*\}$. 

Modern DNN-based models are trained with cross-entropy loss. With such a criterion, the global representation $\mathbf{z}_g$ is trained to minimize the conditional entropy over class label $H(y\mid\mathbf{z}_g)$, which relates to the lower bound of the final cross-entropy loss. This objective is also equivalent to maximizing the mutual information between the features and the class label, as $H(y\mid\mathbf{z}_g)=H(y)-I(y;\mathbf{z}_g)$, where the entropy of $y$ is a constant value. We next decompose the maximum mutual information objective with the following theorem. The proofs for two theorems are in the Appendix A. 

\begin{theorem}
\label{theorem1}
For model $\mathcal{W}$  with the given properties, the mutual information maximization objective decomposes to $I(y;\mathbf{z}_g)=I(y;\mathbf{z}_f) - I(y;\mathbf{z}_f\mid\mathbf{z}_g)$, where maximizing $I(y;\mathbf{z}_f)$ is the classification objective and minimizing $I(y;\mathbf{z}_f\mid\mathbf{z}_g) \ge 0$ is a regularization term.
\end{theorem}

The regularization term $-I(y;\mathbf{z}_f\mid\mathbf{z}_g)$ is to some extent unclear. To interpret how it functions, we next provide an analysis of its optimal solution.

\begin{theorem}
\label{theorem2}
The regularization term is optimized to zero if \textbf{1) constant value solution}: $\mathbf{z}_b$ is a constant value, or \textbf{2) orthogonal subspace solution}: $\mathbf{z}_f$ and $\mathbf{z}_b$ are from different feature subspaces, i.e., $\mathbf{z}_g$ can be equivalently represented by the concatenation of $\mathbf{z}_f$ and $\mathbf{z}_b$.
\end{theorem}

\textbf{Constant value solution.} As ReLU activation maps all negative raw inputs to zero, the only feasible constant value is zero. Otherwise, raw inputs before GAP must stay constant, which is improbable. Our theorems demonstrate that the GAP regularizer suppresses background feature intensity to make their global feature zero, implicitly treating backgrounds as auxiliary outliers, which links backgrounds and available outliers in OE~\cite{hendrycks2018deep}. During training, OE optimizes an auxiliary task that samples from unknown classes should have minimum Maximum SoftMax Probability (MSP), \ie,~predicting uniform probability for known classes on unknown samples. The training objective is defined as:
\begin{align}
    \mathcal{L} = \mathbb{E}_{\left(\mathbf{x},y\right)\sim \mathcal{D}_{in}}\left(H(y;\mathbf{x})\right) + \alpha \mathbb{E}_{\mathbf{x}\sim \mathcal{D}_{oe}}\left(H(u;\mathbf{x})\right),
\end{align}
where $\mathcal{D}_{in}$ represents the distribution for known classes, $\mathcal{D}_{oe}$ is the distribution for known outliers and $u$ is the uniform class distribution, and $\alpha$ is the weight for the auxiliary objective. As is pointed out in~\cite{dhamija2018reducing}, such a regularization objective makes the deep feature from the penultimate layer zero for outliers. 

\textbf{The orthogonal subspace solution.} Under this solution, the model encodes foregrounds and backgrounds as separate feature sets, which leads the model to learn discriminative features for distinguishing between foregrounds and backgrounds. Fore-background discrimination improves model robustness as well as enriches the hidden features, and more diverse features can potentially improve closed-set performance.

\begin{figure}[t]
  \begin{center}
    \includegraphics[width=1\linewidth]{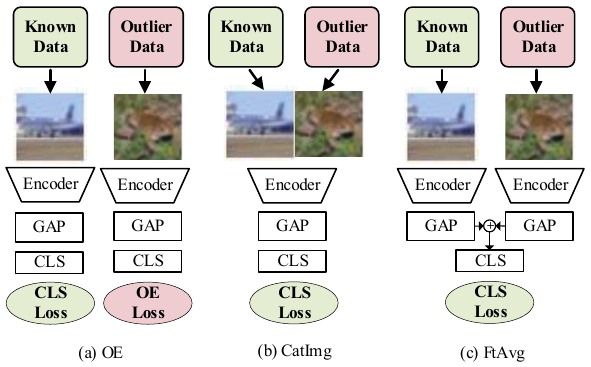}
    \caption{Illustration of three different models compared in the OE experiment. 
    (a) The traditional OE method trains known samples and outliers separately. (b) CatImg concatenates outliers to known samples, serving as constructed image backgrounds. (c) FtAvg inputs known samples and outliers to the backbone separately and restricts them from interacting under simulated GAP.
    }
    \label{fig:oemodels}
    \end{center}
\end{figure}
\subsubsection{Verification experiments}
\label{sec:3.2.2}
To verify our theory and quantitatively evaluate the power of GAP regularizer, we conducted experiments following OE~\cite{hendrycks2018deep}, which is one of the most representative auxiliary data-based methods and has outstanding performance. CIFAR10~\cite{krizhevsky09learning} was used as the known dataset, while LSUN-Fix and ImageNet-Fix~\cite{tack2020csi} were adopted as unknown datasets. LSUN-Fix and ImageNet-Fix contain randomly sampled and resized images from LSUN~\cite{yu2015lsun} and ImageNet~\cite{deng2009imagenet}, respectively. Auxiliary outliers were from TinyImage~\cite{hendrycks2018deep} or CIFAR100~\cite{krizhevsky09learning}. We compared the performance of the following settings depicted in~\figref{fig:oemodels}:

\begin{itemize}
    \item \textbf{Plain}: the closed-set classifier baseline~\cite{hendrycks2016a}, where the auxiliary dataset is not utilized. 
    \item \textbf{OE}: the OE baseline~\cite{hendrycks2018deep}, where the model is regularized to predict a uniform distribution probability across known classes for auxiliary outliers. 
    \item \textbf{CatImg}: an adaptation of GAP regularizer to OE problem. Each in-distribution image (shaped $32\times 32 \times 3$) is concatenated with a random known outlier image on the height dimension to form the $64 \times 32 \times 3$ input.
    \item \textbf{FtAvg}: prevents the foreground and backgrounds from meeting early in the encoder to verify that the regularization functions via GAP. Known class features (serve as $\mathbf{z}_f$) and outlier features (serve as $\mathbf{z}_b$) are mixed by GAP, which follows Eq.~\eqref{eq:decompose_gap} with $\lambda=0.5$. Finally, the classification head inputs the mixed feature and is trained with the original known class label.
\end{itemize}

Besides the mentioned accuracy and AUROC, we evaluated models with three additional metrics: \textbf{TNR@95}, the rate of correctly rejecting unknown samples given a 95\% true positive rate. As AUROC nears one, TNR@95 is more distinguishable in comparison. \textbf{FT-Auc}, the AUROC uses the Euclidean norm of global features as the score function, confirming that the GAP regularizer discovers a constant zero value solution, which suppresses the feature magnitude of unknown samples. \textbf{FT-Cos}: the average cosine similarity between each pair of known and unknown samples. If the model finds an orthogonal subspace solution, FT-Cos approaches zero.

\begin{table*}[t]
  \centering
  \setlength\tabcolsep{6pt}
  \caption{Results for outlier exposure experiments. \revise{The gain or loss values are calculated compared to the plain baseline.}}
    \begin{tabular}{cllllllllll}
    \toprule
    & \multicolumn{1}{r}{} &       & \multicolumn{4}{c}{Test Unknown LSUN-Fix} & \multicolumn{4}{c}{Test Unknown IMGN-Fix} \\
    \cmidrule(r){4-7} \cmidrule(r){8-11}  
    \multicolumn{1}{c}{Outlier} & Method & Accuracy & AUROC   & TNR@95 & FT-Auc & FT-Cos & AUROC   & TNR@95 & FT-Auc & FT-Cos \\
    \midrule
    - & Plain & {95.8} & {89.5} & {45.8} & {66.5} & {43.0} & {89.7} & {48.1} & {66.5} & {43.6} \\ \midrule
    \multirow{3}[1]{*}{TinyImage} & OE & {${95.4}_{\down{0.4\downarrow}}$} & {${98.4}_{\up{8.9\uparrow}}$} & {${96.4}_{\up{50.6\uparrow}}$}  & {${98.6}_{\up{32.1\uparrow}}$}  & {${19.0}_{\down{24.0\downarrow}}$}  & {${97.7}_{\up{8.0\uparrow}}$}  & {${88.6}_{\up{40.5\uparrow}}$}  & {${97.7}_{\up{31.2\uparrow}}$}  & {${18.2}_{\down{25.4\downarrow}}$}  \\
    & CatImg & {${95.7}_{\down{0.1\downarrow}}$}  & {${98.7}_{\up{9.2\uparrow}}$}  & {${93.6}_{\up{47.8\uparrow}}$}  & {${98.9}_{\up{32.4\uparrow}}$}  & {${27.5}_{\down{15.5\downarrow}}$}  & {${97.6}_{\up{7.9\uparrow}}$}  & {${85.8}_{\up{37.7\uparrow}}$}  & {${97.9}_{\up{31.4\uparrow}}$}  & {${26.9}_{\down{16.7\downarrow}}$}  \\
    & FtAvg & {${95.9}_{\up{0.1\uparrow}}$}  & {${98.7}_{\up{9.2\uparrow}}$}  & {${94.4}_{\up{48.6\uparrow}}$}  & {${98.9}_{\up{32.4\uparrow}}$}  & {${25.2}_{\down{17.8\downarrow}}$}  & {${97.7}_{\up{8.0\uparrow}}$}  & {${81.8}_{\up{33.7\uparrow}}$}  & {${97.9}_{\up{31.4\uparrow}}$}  & {${25.4}_{\down{18.2\downarrow}}$}  \\
    \midrule
    \multirow{3}[1]{*}{CIFAR100}  & OE & {${96.1}_{\up{0.3\uparrow}}$}  & {${97.6}_{\up{8.1\uparrow}}$}  & {${88.6}_{\up{42.8\uparrow}}$}  & {${97.4}_{\up{30.9\uparrow}}$}  & {${15.4}_{\down{27.6\downarrow}}$}  & {${97.3}_{\up{7.6\uparrow}}$}  & {${86.9}_{\up{38.8\uparrow}}$}  & {${97.2}_{\up{30.7\uparrow}}$}  & {${16.3}_{\down{27.3\downarrow}}$}  \\
    & CatImg & {${95.8}_{\up{0.0-}}$}  & {${97.4}_{\up{7.9\uparrow}}$}  & {${85.8}_{\up{40.0\uparrow}}$}  & {${97.4}_{\up{30.9\uparrow}}$}  & {${27.5}_{\down{15.5\downarrow}}$}  & {${97.1}_{\up{7.4\uparrow}}$}  & {${83.9}_{\up{35.8\uparrow}}$}  & {${97.1}_{\up{30.6\uparrow}}$}  & {${26.5}_{\down{17.1\downarrow}}$}  \\
    & FtAvg & {${95.7}_{\down{0.1\downarrow}}$}  & {${96.9}_{\up{7.4\uparrow}}$}  & {${81.8}_{\up{36.0\uparrow}}$}  & {${96.9}_{\up{30.4\uparrow}}$}  & {${25.7}_{\down{17.3\downarrow}}$}  & {${96.9}_{\up{7.2\uparrow}}$}  & {${81.8}_{\up{33.7\uparrow}}$}  & {${96.8}_{\up{30.3\uparrow}}$}  & {${25.3}_{\down{18.3\downarrow}}$}  \\
    \bottomrule
    \end{tabular}
  \label{tab:oeexp}%
\end{table*}%

\textbf{GAP regularizers prefer constant zero solution.} 
The performance on FT-Auc in~\tabref{tab:oeexp} shows that the GAP regularizer can achieve even better performance simply with the activation intensity of the global feature. FT-Cos reduction implies that two solutions in Theorem~\ref{theorem2} work together. OE also suppresses and orthogonalizes outlier features, while it shows significantly lower FT-Cos values than GAP regularizers. Thus, we suggest using constant zero activation intensity instead of classification head scores for GAP regularizers.

\textbf{Class-unrelated backgrounds regularize classifier via GAP.} The performance of CatImg and FtAvg shows no significant difference, suggesting GAP effectively regularizes the classifier using background patches. We suppose that local spatial pooling in the encoder has a similar function, explaining a slight overall improvement of CatImg.

\textbf{GAP regularizers show comparable performance without auxiliary optimization tasks.} \tabref{tab:oeexp} reveals that GAP regularizers attain comparable or superior AUROC relative to OE and exhibit robust regularization for known outliers. Meanwhile, OE necessitates manual outlier data specification and a custom loss function, while the GAP regularizer autonomously detects and regulates class-unrelated patches, \ie, well-assumed backgrounds.

\textbf{OE is sensitive to selected outliers.} We additionally used three datasets as auxiliary outliers for training under the OE setting, including DTD~\cite{cimpoi14describing}, LSUN-Fix~\cite{tack2020csi}, and Flower102~\cite{Nilsback2008flower}. DTD is a texture dataset with limited semantic information, LSUN-Fix has minimal overlap with the unknown test dataset, and Flower102 comprises abundant floral data. CIFAR10 or CIFAR100 is used as the known dataset, while the other is the unknown dataset. Besides the mentioned metrics, we used threshold-independent AUPR to evaluate performance based on precision and recall. Results in \tabref{tab:outlier} show that the performance of OE varies significantly depending on the chosen outliers. OE shows impressive performance only when the auxiliary dataset has abundant semantic information. Moreover, the low accuracy in most settings indicates the limitations of OE in improving closed-set performance.

Based on the theoretical analysis and experimental validation, we find that GAP regularizers can use unknown regions within a single image for regularization. Moreover, GAP regularizers exhibit comparable performance to OE and prevent the model from overfitting to selected outliers, thereby ensuring performance stability and reducing the cost.

\begin{table}[t]
  \centering
  \setlength\tabcolsep{9pt}
  \caption{Results for outlier exposure with different datasets served as auxiliary outliers under the OE settings. \revise{The gain or loss values are calculated compared to the baseline without auxiliary outlier.}}
    \begin{tabular}{cllll}
    \toprule
    \multicolumn{1}{c}{Known} & Outlier & Accuracy &TNR@95 & AUROC \\
    \midrule
        \multirow{5}[2]{*}{CIFAR10} & - & {95.8} & {59.7} & {87.9} \\
          & DTD & {${76.1}_{\down{19.7\downarrow}}$} & {${31.4}_{\down{28.3\downarrow}}$} & {${77.4}_{\down{10.5\downarrow}}$}  \\
          & LSUN-Fix & {${81.1}_{\down{14.7\downarrow}}$} & {${33.4}_{\down{26.3\downarrow}}$} & {${77.3}_{\down{10.6\downarrow}}$} \\
          & Flower102 & {${79.7}_{\down{16.1\downarrow}}$} & {${34.8}_{\down{24.9\downarrow}}$} & {${77.6}_{\down{10.3\downarrow}}$} \\
          & TinyImage & {${95.4}_{\down{0.4\downarrow}}$} & {${76.8}_{\up{17.1\uparrow}}$} & {${92.6}_{\up{4.7\uparrow}}$} \\
    \midrule
    \multirow{5}[2]{*}{CIFAR100} & - & {74.5} & {35.3} & {74.7} \\
          & DTD & {${46.7}_{\down{27.8\downarrow}}$} & {${15.7}_{\down{19.6\downarrow}}$} & {${58.7}_{\down{16.0\downarrow}}$} \\
          & LSUN-Fix & {${50.2}_{\down{24.3\downarrow}}$} & {${21.4}_{\down{13.9\downarrow}}$} & {${66.3}_{\down{8.4\downarrow}}$}  \\
          & Flower102 & {${49.7}_{\down{24.8\downarrow}}$} & {${20.7}_{\down{14.6\downarrow}}$} & {${64.5}_{\down{10.2\downarrow}}$}  \\
          & TinyImage & {${75.9}_{\up{1.4\uparrow}}$} & {${39.9}_{\up{4.6\uparrow}}$} & {${78.1}_{\up{3.4\uparrow}}$}  \\
    \bottomrule
    \end{tabular}
  \label{tab:outlier}%
\end{table}%
\section{Background Mix for Open Set Recognition}
\label{sec:method}
Under the common OSR setting, only known class samples and their labels are provided. Neither segmentation annotation nor auxiliary outlier data is available during training. We now propose a solution to apply our findings in practical open set recognition tasks.

\begin{figure}[t]
  \begin{center}
    \includegraphics[width=\linewidth]{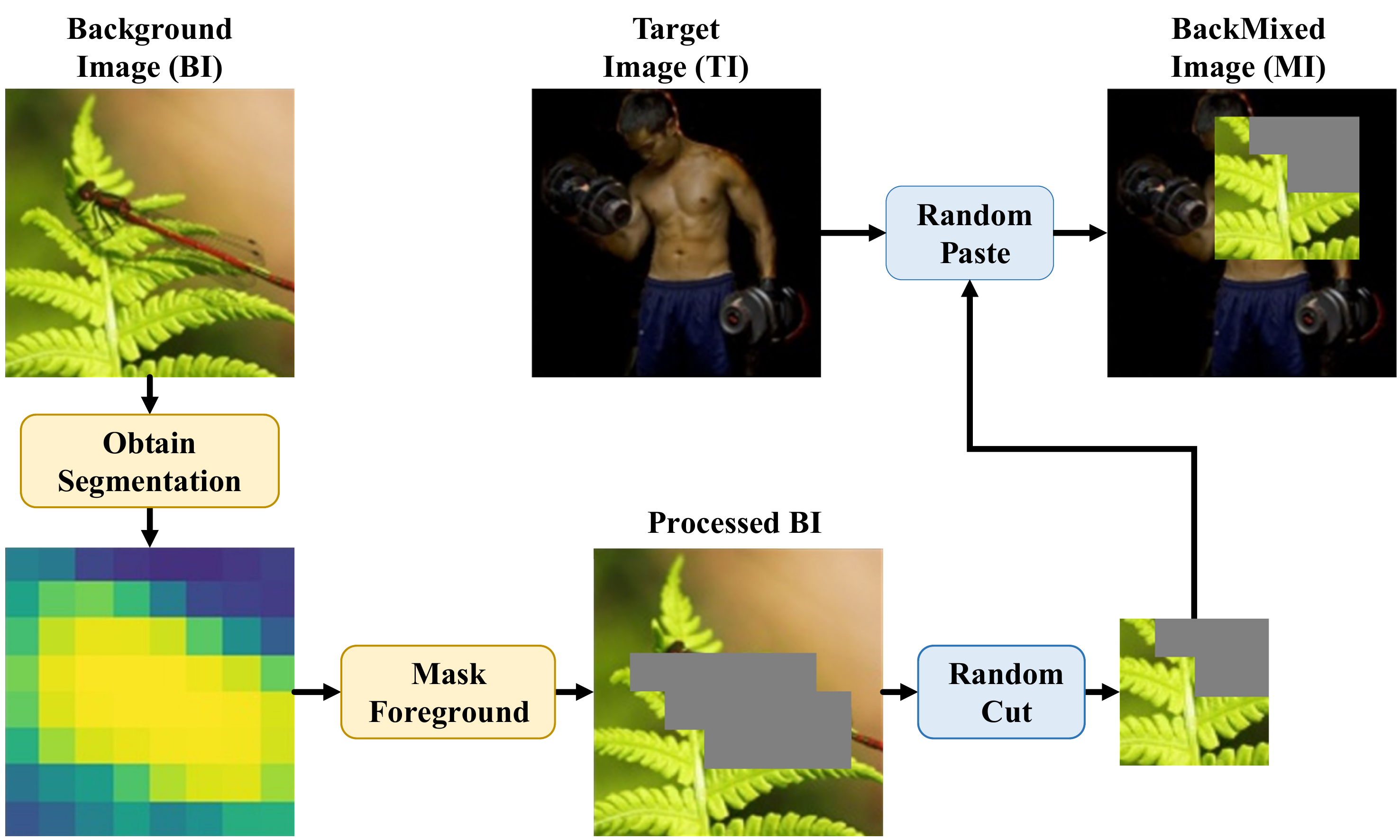}
    \caption{Illustration of BackMix. BackMix first estimates and masks the foreground of the background image, then randomly cuts patches and pastes them on the target image to obtain the mixed image as the training sample.}
    \label{fig:backmix}
    \end{center}
\end{figure}

\subsection{The Proposed Method}
\label{sec:4.1}
We generally follow the cut-and-paste operation in Cutmix. Instead of treating both images equally as Cutmix does, we assign specific roles for the two images to be mixed. One image is designated as the target image (TI), and its label is used for training. The other image is identified as the background image (BI), whose foreground has been masked. Our main idea is to remove multiple foreground instances as well as multiple training objectives.

To mask out the foreground in BI without available annotations, we use Class Activation Maps (CAMs) for approximate estimation. Zhou~\etal~\cite{Zhou_2016_CVPR} employed GAP to generate CAMs, providing an estimation of visual cues for specific categories. The design of CAM leverages the fact that regions relevant to the model's predictions have higher activation values compared to the background. We modify the classification head from \texttt{GAP-Linear-SoftMax} to \texttt{Conv.$1\times 1$-GAP-SoftMax}. Despite their computational equivalence, the latter architecture directly produces CAMs via the $1\times 1$ convolution. We also apply a pixel-wise SoftMax across channels to ensure values remain within the $[0,1]$ range. The channel corresponding to the ground-truth category is then used to estimate the foreground regions.

Formally, suppose $(\mathbf{x}_T,y_T)$ and $(\mathbf{x}_B,y_B)$ are TI and BI, respectively, we generate the mixed sample $(\tilde{\mathbf{x}},\tilde{y})$ by 
\begin{align}
    \tilde{\mathbf{x}} =  \mathbf{M}_b \odot (1-\mathbf{C}_B) \odot \mathbf{x}_B + (1 - \mathbf{M}_b) \odot \mathbf{x}_T, \ \ \tilde{y} = y_T, \label{eq:mixop}
\end{align}
where $\mathbf{M}_b\in \{0,1\}^{H\times W}$ denotes a binary mask of the region where $\mathbf{x}_T$ cuts out its image patches and pastes the patches from $\mathbf{x}_B$, $\mathbf{C}_B \in \{0,1\}^{H\times W}$ denotes the foreground mask, and $\odot$ denotes the element-wise multiplication. Rather than directly using the raw soft estimation from CAM, we sharpen the activation map to generate $\mathbf{C}_B$, where the highest $k$ values are set to one and the rest set to zero. 

The framework of BackMix is illustrated in \figref{fig:backmix} and is detailed in Algorithm~\ref{alg:backmix}. At the very beginning of training, we initialize a bank $\mathbf{C}$ to store the rough foreground estimation generated with CAMs. In the consequent training steps, we randomly pick an image as TI $(\mathbf{x}_T,y_T)$ and another image as BI $(\mathbf{x}_B,y_B)$ within a batch, and then compute the foreground mask $\mathbf{C}_B$ with $\mathbf{C}$ and the mask ratio $k$. Then, we mask out the foreground of $\mathbf{x}_B$ and mix it with $\mathbf{x}_T$ to generate $\Tilde{\mathbf{x}}$, and the value of mix region $\mathbf{M}$ depending on cut size $s$. After that, we update stored mask $\mathbf{C}$ with the CAM $\hat{\mathbf{C}}$ generated by the network. Finally, the mixing process terminates until all images have been served as TI, and the mixed images $\Tilde{\mathbf{X}}$ are used to train the network and update the parameter.

During the initial training steps, CAM may not provide an accurate foreground estimation as the network has yet to learn effective representations. To tackle this, we use a uniform distribution to initialize the mask bank $\mathbf{C}$ and update these masks using an exponential moving average as follows:
\begin{align}
    \mathbf{C}_t=\beta\cdot\hat{\mathbf{C}}_{t}+(1-\beta)\cdot\mathbf{C}_{t-1},
    \label{eq:ema_mask}
\end{align}
where $\mathbf{C}_{t-1}$ and $\mathbf{C}_{t}$ represent the estimated masks obtained in the $t-1$-th and $t$-th training step, respectively. $\hat{\mathbf{C}}_{t}$ denotes the mask calculated only using CAM in $t$-th step and $\beta$ is the exponential decay rate.

As we train the model with only the label of TI, avoiding the effect of foreground objects in the pasted patches is quite important. Unlike the random cut size used in Cutmix, we cut a square region of a fixed size. With a reasonable cut size, \eg, half of the image width or height, TI is guaranteed to dominate the label even if unexpected objects are pasted, which accounts for at most 25\% of the mixed image. One concern is that the masked regions might not provide background information. However, this actually falls into the Cutout if the cut region is fully masked. Such a situation also helps regularize the classifier. Visualizations in Section~\secref{sec:vis} demonstrate that BackMix can effectively estimate the foreground of BI and avoid obscuring the main object in TI.
\begin{algorithm}[t]  
  \caption{Pseudo-code for BackMix}
  \label{alg:backmix}  
  \begin{algorithmic}[0]  
    \Require $\mathcal{D}_{(\mathbf{x},y)}$: Training Set,
      $k$: mask ratio,
      $s$: cut size.
    \Ensure  
      A model $\mathcal{W}$ that can well handle OSR tasks.\\
    $\mathbf{\theta}\leftarrow$ initialize parameters of model $\mathcal{W}$.\\
    $\mathbf{C}\leftarrow$ initialize a CAM bank as soft foreground estimation.
    \Repeat
    \Repeat
      \State Pick a TI $(\mathbf{x}_T,y_T)$ and a BI $(\mathbf{x}_B,y_B)$ within a batch.
      \State Compute $\mathbf{C}_B$ with $k$ and $\mathbf{C}$.
      \State Compute mix region $\mathbf{M}$ with $s$.
      \State Generate sample $\tilde{\mathbf{x}} = \mathbf{M} \odot (1-\mathbf{C}_B) \odot \mathbf{x}_B + (1 - \mathbf{M}) \odot \mathbf{x}_T$.
    \Until{\textit{all images have been selected as TI}}  
      \State Obtain $\hat{\mathbf{C}}$ with respect to $\mathbf{X}$.
      \State Update segmentation $\mathbf{C}\leftarrow \beta\cdot\hat{\mathbf{C}}+(1-\beta)\cdot\mathbf{C}$.
      \State $g_\mathbf{\theta} \leftarrow \nabla\mathbf{\theta} \rm{log}\big(\mathcal{W}_\theta(\tilde{\mathbf{X}})\big)+\rm{log}\big(1-\mathcal{W}_\mathbf{\theta}(\tilde{\mathbf{X}})\big)$.
      \State $\mathbf{\theta} \leftarrow \mathbf{\theta} -\eta g_\mathbf{\theta}$.
    \Until{\textit{convergence}}  
  \end{algorithmic}  
\end{algorithm} 

During the test phase, samples with scores below the threshold are deemed unknown and are rejected, while the remaining samples are classified as known and the model outputs corresponding predictions. The threshold is set to ensure that 95\% of the known samples are correctly classified. As BackMix can be considered as a data augmentation technique, it is applicable for any score function of the original method. For example, when applying it to the SoftMax baseline, the MSP is used as the score function.

\subsection{Discussion}
\textbf{Is it necessary to segment foregrounds precisely?} Generating foreground masks via CAM may not yield precise segmentation, because CAM only highlights the discriminative regions predicted by the model. If the discriminative regions are pasted to TI, the model can be confused, as the true label changes from $y_B$ to $y_T$. Masking out these regions can preserve the well-learned knowledge for class $y_B$. In cases where image backgrounds are mistakenly estimated as foregrounds with high probability, a strategy that masks out only a certain fraction ($k$) of the foregrounds would be beneficial. As long as the real foregrounds have higher confidence, the backgrounds remain and are pasted to TI. After several training steps, these mistakes can be corrected as they are randomly pasted onto all samples. Consequently, our framework can boost the quality of foreground segmentation by itself during training. 

\begin{table}[t]
  \setlength\tabcolsep{3.5pt}
  \centering
  \caption{AUROC score comparison of different OSR methods in unknown detection tasks.}
    \begin{tabular}{llllll}
    \toprule
     Method & SVHN & CIFAR10 & CIFAR+10 & CIFAR+50 & Tiny-IN \\
    \midrule
     OSRCI \cite{neal2018open} & {91.0} & {69.9} & {83.8} & {82.7} & {58.6} \\
     CROSR \cite{yoshihashi2019classification} & {89.9} & {88.3} & {91.2} & {90.5} & {58.9} \\
     C2AE \cite{oza2019c2ae} & {92.2} & {89.5} & {95.5} & {93.7} & {74.8} \\
     CGDL \cite{sun2020conditional} & {93.5} & {90.3} & {95.9} & {95.0} & {76.2} \\
     GDFR \cite{perera2020generative} & {93.5} & {83.1} & {91.5} & {91.3} & {64.7}\\
     PROSER \cite{zhou2021learning} & {94.3} & {89.1} & {96.0} & {95.3} & {69.3} \\
    \midrule
    Plain* & {88.6} & {67.7} & {81.6} & {80.5} & {57.7} \\
    \quad+ BackMix & {${97.0}_{\up{8.4\uparrow}}$} & {${91.3}_{\up{23.6\uparrow}}$} & {${91.9}_{\up{10.3\uparrow}}$} & {${91.6}_{\up{11.1\uparrow}}$} & {${80.4}_{\up{22.7\uparrow}}$} \\
    \midrule
    ARPL \cite{chen2021adversarial} & {95.3} & {89.8} & {91.3} & {90.8} & {76.0} \\
    \quad+ BackMix & {${96.4}_{\up{1.1\uparrow}}$} & {${91.0}_{\up{1.2\uparrow}}$} & {${93.4}_{\up{2.1\uparrow}}$} & {${92.3}_{\up{1.5\uparrow}}$} & {${76.3}_{\up{0.3\uparrow}}$} \\
    \midrule
    CSSR \cite{huang2023cssr} & {96.7} & {90.7} & {91.5} & {90.9} & {80.6} \\
    \quad+ BackMix &{$\textbf{97.7}_{\up{1.0\uparrow}}$}&{$\textbf{94.2}_{\up{3.5\uparrow}}$} & {$\textbf{96.4}_{\up{4.9\uparrow}}$} &{$\textbf{95.7}_{\up{4.8\uparrow}}$} &{$\textbf{83.1}_{\up{2.5\uparrow}}$} \\
    \bottomrule
    \end{tabular}
    \label{tab:unknowndet}%
\end{table}%

\textbf{Is it necessary to mix the labels?} Although we mix two different images, it is not necessary to mix two image labels. The most discriminative regions for BI are masked out and are supposed to make few differences. Note that it is not necessary to paste the BI to the background regions of TI carefully. Pasting BI on foreground regions of TI functions like Cutout, which simulates occlusions for training images. In Section~\secref{sec:exp}, we empirically show that mixing labels of two images could limit the OSR performance.

\section{Experiments}
\label{sec:exp}
\textbf{{Implementation details.}}
We set the fixed cut size to $0.5\times$ the height and width of the image. The exponential decay rate $\beta$ for segmentation masks was set to 0.1. We set $k = 0.25$, \ie,~25\% pixels with the highest segmentation scores are masked out for BI. We trained WideResNet40-4~\cite{zagoruyko2016wide} on small-scale datasets (\eg~CIFAR10) and ResNet18~\cite{he2016deep} on ImageNet30 and Tiny-ImageNet, setting the batch size to 128 and the learning rate to 0.1, with a cosine annealing learning rate scheduler and an SGD optimizer. 

\subsection{Comparison with OSR Methods}
\label{sec:5.1}
 To verify the effectiveness of the proposed method, we applied BackMix to the baseline strategy MSP (Plain*)~\cite{hendrycks2016a} and also integrated it to \sArt~methods ARPL~\cite{chen2021adversarial} and CSSR~\cite{huang2023cssr}. Notice that to ensure a fair comparison, only simple data augmentation (\eg~RandomHorizontalFlip and RandomCrop) used in the original method was retained, and all experiments were conducted under the original settings.

\subsubsection{Unknown detection}
\label{sec:5.1.1}
For the OSR task, we first followed the setting from \cite{neal2018open} to conduct the unknown detection experiments. Five standard benchmarks were adopted in the experiments, including 
CIFAR10~\cite{krizhevsky09learning}, SVHN~\cite{netzer2011reading}, CIFAR+10, CIFAR+50 and Tiny-ImageNet~\cite{pouransari2014tiny}.

\begin{table}[t]
  \centering
  \setlength\tabcolsep{8pt}
  \caption{Open set recognition performance with CIFAR10 as known and various datasets as unknown.}
    \begin{tabular}{lllll}
    \toprule
    Method & IMGN-C & IMGN-R & LSUN-C & LSUN-R \\
    \midrule
    CROSR~\cite{yoshihashi2019classification} & {72.1} & {73.5} & {72.0} & {74.9} \\
    GDFR~\cite{perera2020generative} & {75.7} & {79.2} & {75.1} & {80.5} \\
    C2AE~\cite{oza2019c2ae} & {83.7} & {82.6} & {78.3} & {80.1} \\
    CGDL~\cite{sun2020conditional} & {84.0} & {83.2} & {80.6} & {81.2} \\
    PROSER~\cite{zhou2021learning} & {84.9} & {82.4} & {86.7} & {85.6} \\
    ConOSR~\cite{xu2023contra} & {89.1} & {84.3} & {91.2} & {88.1} \\
    \midrule
    Plain* & {63.9} & {65.3} & {64.2} & {64.7} \\
    \quad+ BackMix  & {${92.6}_{\up{28.7\uparrow}}$}  & {${90.4}_{\up{25.1\uparrow}}$}  & {${92.6}_{\up{28.4\uparrow}}$} & {${93.3}_{\up{28.6\uparrow}}$}  \\
    \midrule
    ARPL~\cite{chen2021adversarial} & {80.6} & {82.5} & {85.3} & {82.7} \\
    \quad+ BackMix & {${92.3}_{\up{11.7\uparrow}}$} &{${91.3}_{\up{8.8\uparrow}}$} & {${92.9}_{\up{7.6\uparrow}}$} & {${94.2}_{\up{11.5\uparrow}}$} \\
    \midrule
    CSSR~\cite{huang2023cssr} & {88.3} & {89.5} & {92.2} & {90.4} \\
    \quad+ BackMix & {$\textbf{93.7}_{\up{5.4\uparrow}}$} &{$\textbf{93.0}_{\up{3.5\uparrow}}$} & {$\textbf{94.7}_{\up{2.5\uparrow}}$} & {$\textbf{94.8}_{\up{4.4\uparrow}}$} \\
    \bottomrule
    \end{tabular}
  \label{tab:osr}%
\end{table}
For ten-class datasets CIFAR10 and SVHN, we randomly selected six classes as known classes to appear during training, and the remaining four classes as unknown classes for testing. For CIFAR+{\textit{N}} datasets, we selected four non-animal classes from CIFAR10 as known classes, while using {\textit{N}} animal classes from CIFAR100 as unknown classes and set {{\textit{N}}=10} and {{\textit{N}}=50} to test performance at the different scenarios. For the large-scale and more challenging dataset Tiny-ImageNet (Tiny-IN), we selected 20 classes as known and the remaining 180 classes as unknown. Following the standard evaluation protocol, we adopted the threshold-independent metric AUROC, and all the reported results were the average of five trials.  

\textbf{BackMix significantly enhances the open set performance of various methods and surpasses the~\sArt~by removing the fore-background priors.} We compared the proposed method with classic and advanced OSR methods. The performance values of other methods were from \cite{huang2023cssr,oza2019c2ae,perera2020generative,sun2020conditional,yoshihashi2019classification,zhou2021learning} or reproduced with the official code under our settings. Results in Table~\ref{tab:unknowndet} show that BackMix significantly improves the performance of the plain SoftMax with little cost, making it exceed many complex methods, especially in the CIFAR10 (+23.6\%) and Tiny-ImageNet (+22.7\%) experiments. In addition, by applying our method to \sArt~methods, ARPL, and CSSR, they obtained consistent performance improvement in all settings and achieved new \sArt~performance. Therefore, we demonstrated the effectiveness of the proposed method design and its applicability.

\begin{table*}
  \centering
  \setlength\tabcolsep{10pt}
  \caption{Comparision for distinguishing known dataset CIFAR10 from near OOD dataset CIFAR100 and far OOD dataset SVHN. }
    \begin{tabular}{lllllllll}
    \toprule
    \multirow{2}[1]{*}{Method} & \multicolumn{4}{c}{In:CIFAR10 / Out:CIFAR100} & \multicolumn{4}{c}{In:CIFAR10 / Out:SVHN} \\
    \cmidrule(r){2-5} \cmidrule(r){6-9} & DTACC & AUROC & AUIN  & AUOUT & DTACC & AUROC & AUIN  & AUOUT \\
    \midrule
    GCPL~\cite{yang2022cpl}  & 80.2  & 86.4  & 86.6  & 84.1  & 86.1  & 91.3  & 86.6  & 94.8 \\
    RPL~\cite{chen2020learning}   & 80.6  & 87.1  & 88.8  & 83.8  & 87.1  & 92.0    & 89.6  & 95.1 \\
    CSI~\cite{tack2020csi} & 84.4 & 91.6 & 92.5 & 90.0 & 92.8 & 97.9 & 96.2 & 99.0 \\
    OpenGAN~\cite{kong2021opengan} & 84.2 & 89.7 & 87.7 & 89.6 & 92.1 & 95.9 & 93.4 & 97.1\\
    \midrule
    Plain* & 79.8 & 86.3 & 88.4 & 82.5 & 86.4 & 90.6 & 88.3 & 93.6 \\
    \quad+ BackMix & {${84.9}_{\up{5.1\uparrow}}$} & {${91.3}_{\up{5.0\uparrow}}$} & {${93.0}_{\up{4.6\uparrow}}$} & {${88.1}_{\up{5.6\uparrow}}$} & {${88.5}_{\up{2.1\uparrow}}$} & {${94.1}_{\up{3.5\uparrow}}$} & {${93.5}_{\up{5.2\uparrow}}$} & {${97.5}_{\up{3.9\uparrow}}$} \\
    \midrule
    ARPL~\cite{chen2021adversarial} & 80.8 & 88.2 & 90.4 & 84.4 & 82.8 & 90.5 & 84.6 & 95.3 \\
    \quad+ BackMix & {${84.0}_{\up{3.2\uparrow}}$} & {${91.1}_{\up{2.9\uparrow}}$} & {${92.1}_{\up{1.7\uparrow}}$} & {${89.0}_{\up{4.6\uparrow}}$} & {${94.9}_{\up{12.1\uparrow}}$} & {${98.5}_{\up{8.0\uparrow}}$} & {${97.6}_{\up{13.0\uparrow}}$} & {${99.1}_{\up{3.8\uparrow}}$} \\
    \midrule
    CSSR~\cite{huang2023cssr} & 83.1 & 90.3 & 91.3 & 87.8 & 94.1 & 98.1 & 97.1 & 98.2 \\
    \quad+ BackMix &{$\textbf{86.3}_{\up{3.2\uparrow}}$}  & {$\textbf{93.0}_{\up{2.7\uparrow}}$}  & {$\textbf{93.7}_{\up{2.4\uparrow}}$}  & {$\textbf{91.7}_{\up{3.9\uparrow}}$}  & {$\textbf{96.4}_{\up{2.3\uparrow}}$}  & {$\textbf{99.2}_{\up{1.1\uparrow}}$}  & {$\textbf{98.4}_{\up{1.3\uparrow}}$}  & {$\textbf{99.6}_{\up{1.4\uparrow}}$} \\
    \bottomrule
    \end{tabular}
  \label{tab:ood}%
\end{table*}%

\subsubsection{Open set classification}
\label{sec:5.1.2}
We then followed a common experimental setup from~\cite{yoshihashi2019classification} to test the performance when introducing unknown samples from other datasets. The models were trained on CIFAR10, while in the test phase, the samples from ImageNet~\cite{russakovsky2015imagenet} and LSUN~\cite{yu2015lsun} were used as unknowns. These two datasets are cropped or resized so that their image size can remain the same as known samples, which form ImageNet-Crop (IMGN-C), ImageNet-Resize (IMGN-R), LSUN-Crop (LSUN-C), and LSUN-Resize (LSUN-R). For a fair comparison, we used the release version of the four datasets from~\cite{liang2018enhancing}. The performance was evaluated by macro-averaged F1-scores in 11 classes (including 10 known classes and 1 unknown class). 

\textbf{BackMix establishes impressive capability in classifying known samples and distinguishing unknown samples.} We reported results in Table~\ref{tab:osr} and the values other than ours are taken from~\cite{yang2022cpl,sun2020conditional,zhou2021learning,tack2020csi} or reproduced using the official code under our settings. We can observe that BackMix makes the plain baseline exceed many recent complex methods and further improves the CSSR~\cite{huang2023cssr} to outperform existing \sArt~OSR methods by a significant margin.

\subsubsection{Out-of-Distribution detection}
\label{sec:5.1.3}
Considering the unknown classes from different datasets, we also followed the setting of~\cite{chen2021adversarial} to carry out the out-of-distribution detection experiment. For the out-of-distribution (OOD) detection task, in addition to the AUROC and AUPR, we used the DTACC following~\cite{chen2021adversarial}, which calculates the maximum known or unknown classification accuracy a model can achieve over all possible decision thresholds. The metric AUPR becomes AUIN (or AUOUT) if known (or unknown) samples are specified as the positive class.

\begin{table}
  \centering
  \setlength\tabcolsep{9pt}
  \caption{Comparison for closed-set classification and open set detection performance on split ImageNet30.}
    \begin{tabular}{lllllllll}
    \toprule
    Augmentation & Accuracy & AUROC & AUIN  & AUOUT \\
    \midrule
    Plain*  & {94.9} & {89.9} & {86.4} & {92.9}\\
    Cutout~\cite{devries2017improved} & {${95.7}_{\up{0.8\uparrow}}$} & {${90.4}_{\up{0.5\uparrow}}$} & {${86.7}_{\up{0.3\uparrow}}$} & {${92.9}_{\up{0.0-}}$}\\
    Mixup~\cite{zhang2017mixup} & {${95.9}_{\up{1.0\uparrow}}$} & {${89.8}_{\down{0.1\downarrow}}$} & {${80.8}_{\down{5.6\downarrow}}$} & {${94.0}_{\up{1.1\uparrow}}$}\\
    Cutmix~\cite{yun2019cutmix} & {${96.8}_{\up{1.9\uparrow}}$} & {${89.7}_{\down{0.2\downarrow}}$} & {${76.7}_{\down{9.7\downarrow}}$} & {${93.5}_{\up{0.6\uparrow}}$}\\
    \midrule
    BackMix & {$\textbf{97.2}_{\up{2.3\uparrow}}$} & {$\textbf{91.3}_{\up{1.4\uparrow}}$} & {$\textbf{87.7}_{\up{1.3\uparrow}}$} & {$\textbf{94.1}_{\up{1.2\uparrow}}$}\\
    \bottomrule
    \end{tabular}
  \label{tab:cmp_aug}%
\end{table}%

\textbf{BackMix improves performance on detecting unknown samples from datasets that are either slightly or significantly different from the known dataset.} Results in Table \ref{tab:ood} indicate that applying BackMix on different methods significantly improves their performance of OOD detection tasks and reached higher results compared with \sArt~methods. Concretely, BackMix enhanced ARPL by up to 13.0\% on AUIN and CSSR by up to 3.9\% on AUOUT. Note that the improvement brought by BackMix on CSSR is less obvious compared to other baselines, which is probably because CSSR already had outstanding performance. \revise{Besides, we applied BackMix on the plain baseline with various post-processing methods as score functions in the Appendix~\ref{A_3_2}. BackMix consistently improves performance on both prediction-based and feature-based score functions.}

\subsection{Comparison with Data Augmentation Techniques}
\label{sec:5.2}
Since the BackMix can be seen as a data augmentation technique, we compared it with three commonly used strong techniques—Cutout~\cite{devries2017improved}, Mixup~\cite{zhang2017mixup}, and Cutmix~\cite{yun2019cutmix}—using the plain SoftMax strategy. As discussed in Section \secref{sec:2.2}, these techniques inherently mix backgrounds.

To evaluate the closed-set and open set performance, we adopted several metrics including closed-set accuracy, AUROC, AUIN (or AUOUT), and DTACC as used in previous experiments. All the above metrics are threshold-free. Following the standard setting in \secref{sec:5.1.1}, we conducted experiments on split ImageNet30~\cite{hendrycks2019using}, which contains 30 classes with 1300 training images and 100 test images per class, and the image resolution is 224$\times$224. We took the first 10 classes as known classes and the rest 20 as unknown classes during test. 

\textbf{BackMix is more suitable for the OSR task compared to other data augmentation techniques.} Results in \tabref{tab:cmp_aug} indicate that BackMix obtains a significant improvement in both closed-set and open set scenarios. The compared methods enhanced the closed-set accuracy but obtained limited improvement or even performance drop on open set metrics, which infers that mixing labels may help improve closed-set performance but can limit open set performance. The proposed BackMix used hard labels and avoided multiple objects in foregrounds, keeping both closed and open set performance.
\begin{table}[b]
  \centering
  \setlength\tabcolsep{12pt}
  \caption{Closed-set classification accuracy performance comparison on the CIFAR10 dataset.}
    \begin{tabular}{ll}
    \toprule
    Method  & Accuracy \\
    \midrule
    CROSR \cite{yoshihashi2019classification} & 94.0 \\
    CGDL \cite{sun2020conditional}   & 91.2 \\
    GCPL \cite{yang2022cpl}  & 93.3 \\
    \midrule
    Plain* & 94.0 \\
    \quad+ BackMix & {${95.1}_{\up{1.1\uparrow}}$} \\
    \midrule
    ARPL \cite{chen2021adversarial} & 92.7 \\
    \quad+ BackMix & {${93.2}_{\up{0.5\uparrow}}$} \\
    \midrule
    CSSR \cite{huang2023cssr} & 94.2 \\
    \quad+ BackMix & {$\textbf{95.6}_{\up{1.4\uparrow}}$} \\
    \bottomrule
    \end{tabular}
  \label{tab:close_performance}%
\end{table}%

\subsection{Further Analysis}
\label{sec:5.3}
\subsubsection{Closed-set classification performance}
\label{sec:5.3.1}
For the OSR task, accurate classification of known classes is also very important. Therefore, we compared the closed-set performance of different methods on CIFAR10. To test the effectiveness of BackMix in improving classification performance, we applied it to three baselines: plain SoftMax, ARPL, and CSSR.

\begin{figure}[t]
\subfigure[]{  
        \label{fig:sub1}  
        \includegraphics[width=0.465\linewidth]{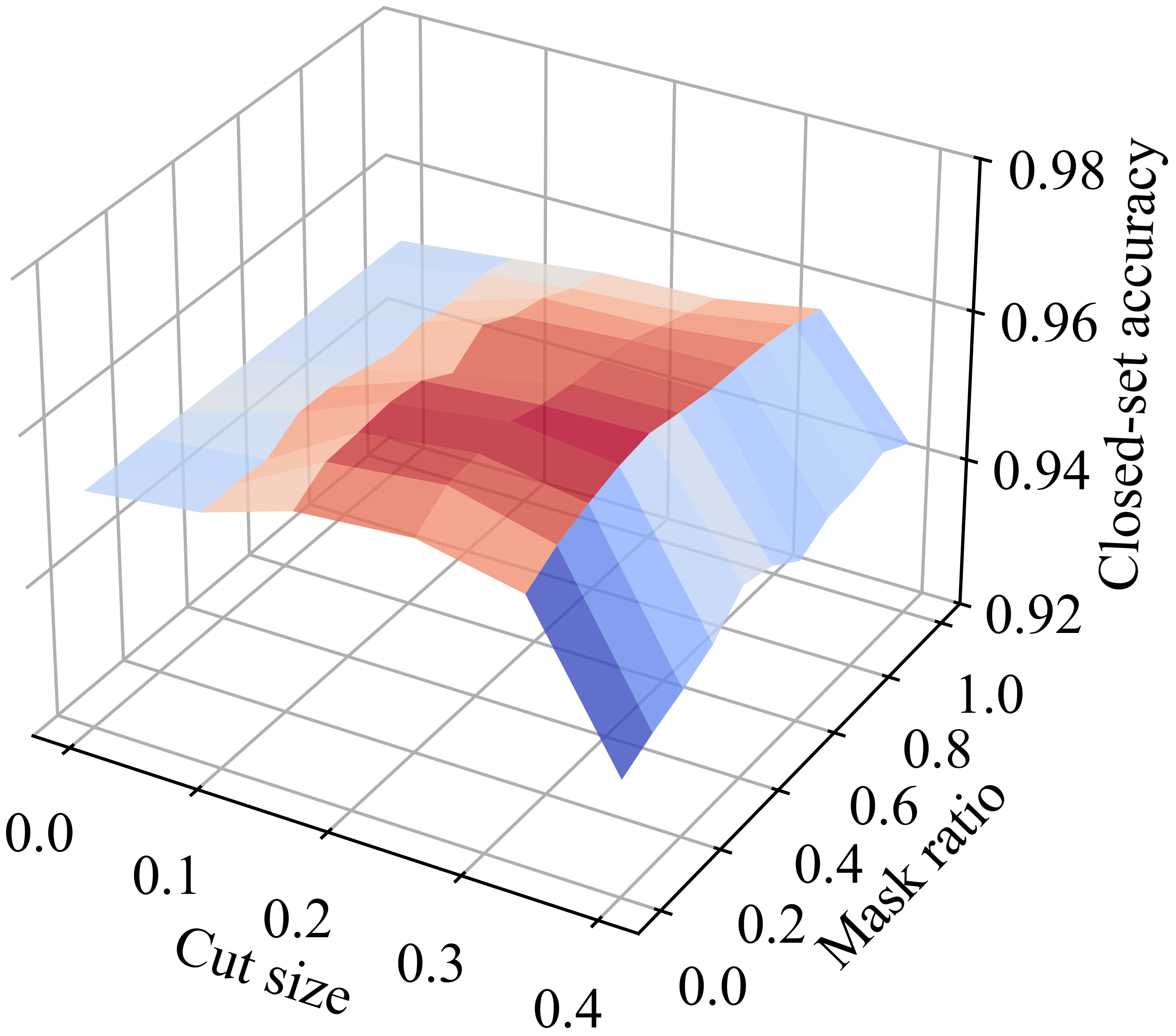}
    }
    \hfill
    \subfigure[]{  
        \label{fig:sub2}  
        \includegraphics[width=0.465\linewidth]{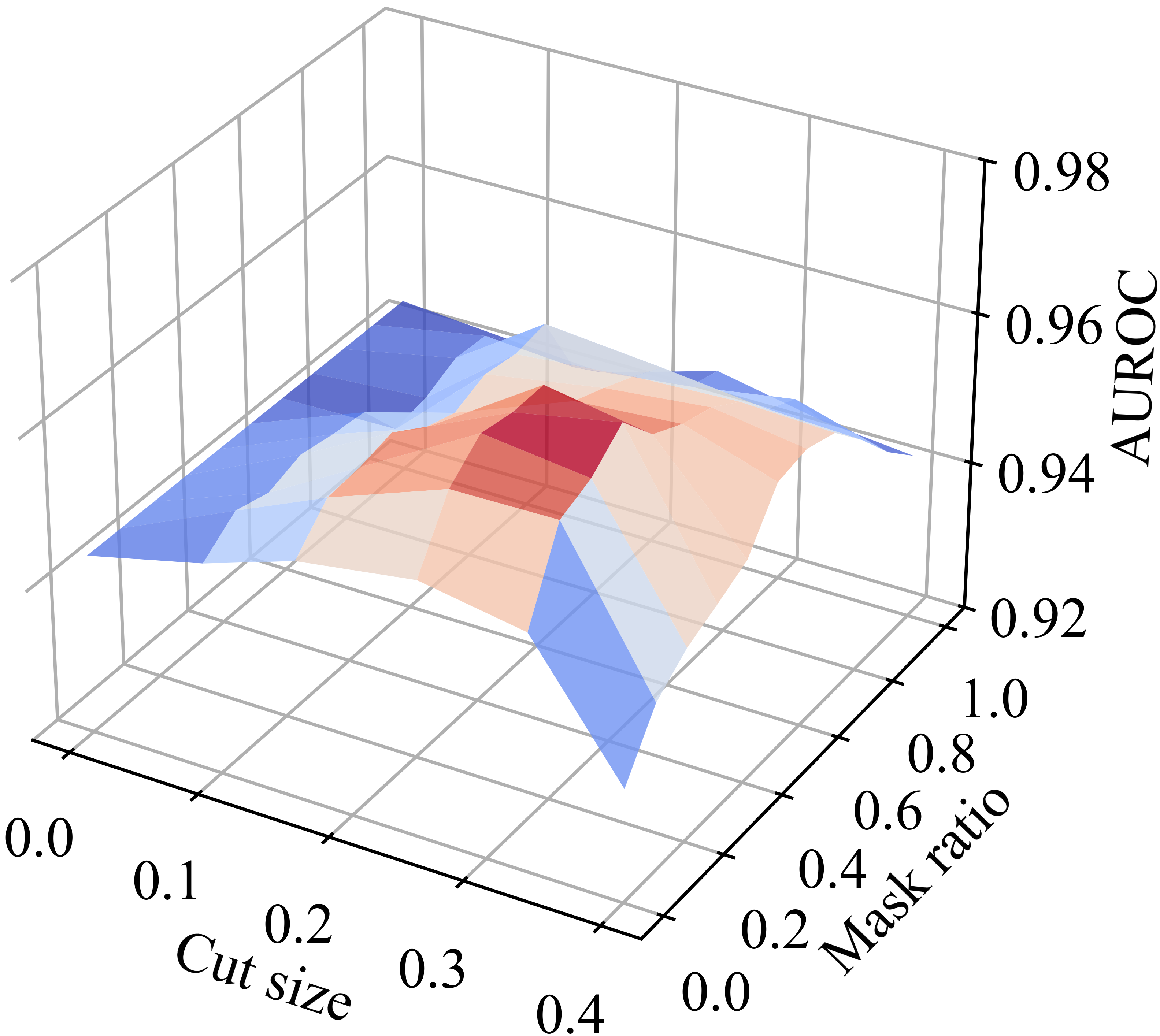}  
    }  
    \caption{\revise{Parameter Analysis for BackMix, where (a) presents the closed-set classification accuracy, (b) presents the AUROC with varying cut size $s$ and mask ratio $k$. The values of open set metric AUROC are averaged on the six unknown datasets.}}
    \label{fig:parameter}
\end{figure}

Results in Table \ref{tab:close_performance} show that applying BackMix on different baselines obtains consistent improvement in the closed-set performance, and previous experiments show that BackMix improves the open set performance concurrently, which indicates BackMix is suitable for the OSR task.
\subsubsection{\revise{Is BackMix sensitive to hyperparameters}}
\label{sec:5.3.2}
\revise{We performed parameter analysis on the cut size $s$ and mask ratio $k$ here. Cut size $s$ represents the ratio of the cutting area to the whole image area, \eg, a cut box sized $0.5\times 0.5$ corresponds to $s=0.25$. In this experiment, ten classes from CIFAR10 were used as known, while six datasets, including ImageNet-Crop, ImageNet-Resize, ImageNet-Fix, LSUN-Crop, LSUN-Resize, and LSUN-Fix, were treated as unknown. We used the preprocessed version in \cite{liang2018enhancing,tack2020csi}. We varied the cut size $s$ from 0 to 0.4 and the mask ratio $k$ from 0 to 1. It should be noted that BackMix converts to Cutout with $k=1$, where BI is processed to be pure grey, and with $k=0$, we do not mask anything in BI at all.}

\revise{\textbf{BackMix maintains stable performance and is not sensitive to hyperparameters except for extreme value cases.} We plotted closed-set accuracy and AUROC results in \figref{fig:parameter}. Proper values (0.2-0.33) for cut size as well as mask ratio boost both closed and open set performance. For the mask ratio $k$, masking out a proper fraction of possible foregrounds does improve open set performance. It is also interesting to find that though the closed-set performance for $k=1$ (Cutout) raises, the open set performance drops. Therefore, we emphasize that mixing backgrounds is important for OSR while avoiding multiple foregrounds is essential for preserving closed-set performance. This meets the conclusion of the empirical study in \secref{sec:3.1}. Additionally, we also demonstrated the robustness of BackMix in background selection in the Appendix~\ref{A_3_1}.}

\subsubsection{\revise{BackMix on large-scale pretrained models}}
\label{sec:5.3.3}
\revise{\textbf{BackMix further improves pretrained model performance in the few-shot finetuning stage.} We used CIFAR10 as the known dataset and CIFAR100 as the unknown dataset to evaluate the performance of model on the few-shot OOD detection task. BackMix was applied to the few-shot finetuning methods CoOp~\cite{zhou2022coop} and LoCoOp~\cite{miyai2023locoop} based on large-scale pretrained model CLIP~\cite{radford2021clip}. Results in~\tabref{tab:ptm} show that BackMix leverages the powerful representational capacity of the large-scale pretrained model and further enhances both the closed-set and open set performance of the model even with only 1 sample from each known class.}
\begin{table}[t]
    \centering
  \setlength\tabcolsep{2.5pt}
    \caption{\revise{OOD detection performance of BackMix in the finetuning stage. We used CIFAR10 as the in-distribution dataset and CIFAR100 as the OOD dataset.}}
    \begin{tabular}{lllllll}
    \toprule
    \multirow{2}[1]{*}{Method} & \multicolumn{2}{c}{1-shot} & \multicolumn{2}{c}{4-shot} & \multicolumn{2}{c}{16-shot} \\\cmidrule(r){2-3} \cmidrule(r){4-5} \cmidrule(r){6-7}
    & Accuracy & AUROC & Accuracy & AUROC & Accuracy & AUROC \\ \midrule
    CoOp~\cite{zhou2022coop} & 89.8 & 91.6 & 90.6 & 91.5 & 91.2 & 91.1\\
    \quad+BackMix & {${90.7}_{\up{0.9\uparrow}}$} & {${92.1}_{\up{0.5\uparrow}}$} & {${91.3}_{\up{0.7\uparrow}}$} &  {${92.1}_{\up{0.6\uparrow}}$} &  {${91.7}_{\up{0.5\uparrow}}$} &  {${91.6}_{\up{0.5\uparrow}}$}\\
    \midrule
    LoCoOp~\cite{miyai2023locoop} & 89.6 & 91.2 & 89.8 & 91.4 & 91.4 & 90.4\\
    \quad+BackMix & {${90.7}_{\up{1.1\uparrow}}$} & {${91.6}_{\up{0.4\uparrow}}$} & {${90.9}_{\up{1.1\uparrow}}$} & {${91.9}_{\up{0.5\uparrow}}$} & {${91.7}_{\up{0.3\uparrow}}$} & {${91.0}_{\up{0.6\uparrow}}$}\\
    \bottomrule
    \end{tabular}
    \label{tab:ptm}
\end{table}

\begin{figure}[b]
  \begin{center}
    \includegraphics[width=\linewidth]{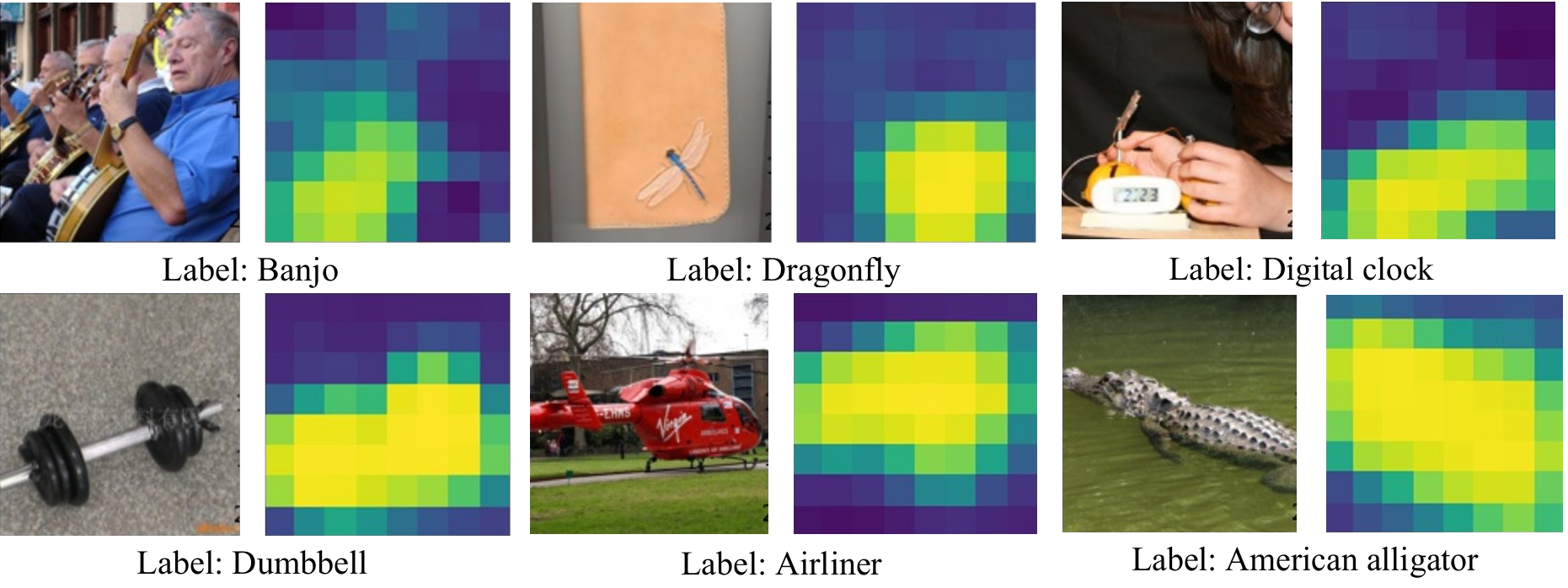}
    \caption{Examples of the estimated foreground masks, and labels have been annotated below the corresponding image. The rough segmentation using CAM can effectively estimate the foreground.}
    \label{fig:foremasks}
    \end{center}
\end{figure}

\begin{figure*}[t]
  \begin{center}
    \includegraphics[width=1.\linewidth]{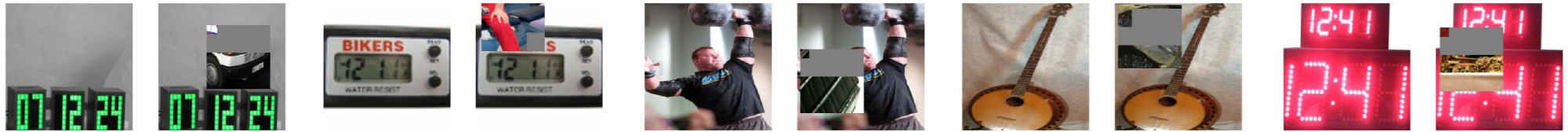}
    \caption{Examples of the BackMix processed images. The pasted background patches contain almost no foreground objects from another image. By setting a reasonable cut size, we can ensure that the processed training samples retain sufficient information about the classification object in the target image.}
    \label{fig:mixeds}
    \end{center}
\end{figure*}

\subsubsection{Visualizations}
\label{sec:vis}
\textbf{BackMix estimates the foreground regions accurately without additional annotations.}~We provide examples of the estimated foreground masks from experiments on ImageNet30. As shown in \figref{fig:foremasks}, the estimations are accurate enough to mask out most of the foreground regions, verifying the effectiveness and feasibility of the BackMix.

\textbf{BackMix prevents the performance degradation caused by foreground occlusion.}~To better illustrate the effectiveness of our method in image processing, we provide some processed samples in~\figref{fig:mixeds}. The cut portions from the BI do not contain significant foreground objects, which avoids misleading the model after being pasted to TI. Due to the setting of proper mixing ratio, our method preserves sufficient classification information for the model to make judgments, regardless of whether the majority or minority of the region of interest is present. Therefore, it avoids situations where key areas are occluded and cannot be accurately classified.

\section{Conclusions}\label{sec:conclusion}
In this paper, we discuss open set recognition from a new perspective of fore-background priors. We explore the role of fore-background priors and provide insights into how classifiers model the backgrounds in OSR. Empirical and theoretical analyses show that the underlying fore-background priors have negative impacts on OSR performance while removing these priors can enhance the performance. More importantly, class-unrelated backgrounds serve as auxiliary known outliers and provide extra regularization via global average pooling. Inspired by the insights, we design a new BackMix method that mixes the foreground of an image with the backgrounds of different images. The proposed method is simple to implement, requires no segmentation annotations or priors, and can seamlessly integrate into the learning process of almost all of the existing methods. Extensive experiments show that our method can significantly enhance state-of-the-art open set recognition methods and show clear advantages over existing data augmentation methods. 

\appendices
\section{Theoretical Proofs}
\label{A_1}
\noindent\textbf{Theorem 1.} \textit{For model $\mathcal{W}$  with the given properties, the mutual information maximization objective decomposes to $I(y;\mathbf{z}_g)=I(y;\mathbf{z}_f) - I(y;\mathbf{z}_f|\mathbf{z}_g)$, where maximizing $I(y;\mathbf{z}_f)$ is the classification objective and minimizing $I(y;\mathbf{z}_f|\mathbf{z}_g) \ge 0$ is a regularization term.}

\begin{proof}
We obtain the proof with data processing inequality. First of all, we have the following equations 
\begin{align}
I(y;\mathbf{z}_g,\mathbf{z}_f)&=I(y;\mathbf{z}_g | \mathbf{z}_f) + I(y;\mathbf{z}_f)\notag \\&= I(y;\mathbf{z}_f | \mathbf{z}_g) + I(y;\mathbf{z}_g). 
\label{eq:mutal_info}
\end{align}
Given foreground feature $\mathbf{z}_f$, global feature $\mathbf{z}_g$ depends only on background feature $\mathbf{z}_b$ (Property \up{1}). Also because background feature $\mathbf{z}_b$ is independent from class label $y$ (Property \up{2}), we have $I(y;\mathbf{z}_g | \mathbf{z}_f)=0$. Therefore, Eq.~\eqref{eq:mutal_info} can be re-organized to $I(y;\mathbf{z}_g)=I(y;\mathbf{z}_f)-I(y;\mathbf{z}_f|\mathbf{z}_g)$.
\end{proof}

\noindent\textbf{Theorem 2.} \textit{
The regularization term is optimized to zero if \textbf{1) constant value solution}: $\mathbf{z}_b$ is a constant value, or \textbf{2) orthogonal subspace solution}: $\mathbf{z}_f$ and $\mathbf{z}_b$ are from different feature subspaces, i.e., $\mathbf{z}_g$ can be equivalently represented by the concatenation of $\mathbf{z}_f$ and $\mathbf{z}_b$.}

\begin{proof}
(a) Let $\mathbf{z}_b$ be the constant, which is no longer a random variable. Thus, $I(y;\mathbf{z}_g)=I(y;\mathbf{z}_f+\mathbf{c})=I(y;\mathbf{z}_f)$, \ie, the extra regularization term is zero. (b) If $\mathbf{z}_f$ and $\mathbf{z}_b$ are from different feature subspaces, the probability density satisfies $p(\mathbf{z}_g)=p(\mathbf{z}_f) p(\mathbf{z}_b)$, then
\begin{align}
p(\mathbf{z}_f|\mathbf{z}_g) p(y|\mathbf{z}_g) &=\frac{p(\mathbf{z}_f,\mathbf{z}_g)p(y,\mathbf{z}_g)}{p(\mathbf{z}_g)p(\mathbf{z}_g)}\notag \\
&=\frac{[p(\mathbf{z}_g|\mathbf{z}_f)p(\mathbf{z}_f)]p(y,\mathbf{z}_f,\mathbf{z}_b)}{p(\mathbf{z}_f)p(\mathbf{z}_b)p(\mathbf{z}_g)}\notag \\&=\frac{p(y,\mathbf{z}_f) p(\mathbf{z}_b)}{p(\mathbf{z}_g)},\label{eq:3}\\
p(y,\mathbf{z}_f|\mathbf{z}_g) &=\frac{p(y,\mathbf{z}_f,\mathbf{z}_g)}{p(\mathbf{z}_g)}= \frac{p(y,\mathbf{z}_g|\mathbf{z}_f)p(\mathbf{z}_f)}{p(\mathbf{z}_g)}\notag\\&= \frac{p(y|\mathbf{z}_f)p(\mathbf{z}_b)p(\mathbf{z}_f)}{p(\mathbf{z}_g)}= \frac{p(y,\mathbf{z}_f) p(\mathbf{z}_b)}{p(\mathbf{z}_g)}. \label{eq:4}
\end{align}
As Eq.~\eqref{eq:3} equals to Eq.~\eqref{eq:4}, we have $p(y,\mathbf{z}_f|\mathbf{z}_g) = p(\mathbf{z}_f|\mathbf{z}_g) p(y|\mathbf{z}_g)$, implying random variables $y$ and $\mathbf{z}_f$ are independent conditioned on $\mathbf{z}_g$. Therefore, we have $I(y;\mathbf{z}_f|\mathbf{z}_g)=0$.
\end{proof}

\section{Verification Experiments Details}
\label{app_exp_details}
\subsection{Experimental Setups for Synthesized Dataset}
\label{A_2_1}
\textbf{Dataset construction.} We chose 18 classes from COCO~\cite{cocods} in total. Specifically, 12 classes were selected as known classes: airplane, bicycle, bird, boat, bottle, bus, car, dog, horse, TV, motorcycle, and person, while 6 classes were selected to serve as unknown classes in the test: sheep, cow, elephant, bench, toilet, and cat. For each class, we cropped at most 2,000 instances according to annotated bounding boxes. To preserve spaces for backgrounds, the crop boxes were given a margin of 100 pixels to the object bounding boxes. When cropping an instance from an image, we cropped the corresponding segmentation mask simultaneously to operate on foregrounds and backgrounds during training.
\begin{table}[t]
    \centering
    \caption{Comparison of using different background selection strategies in the task of unknown detection.}
    \setlength\tabcolsep{2.5pt}
    \begin{tabular}{ccccccc}
         \toprule
         \multirow{2}[1]{*}{Methods} & \multicolumn{2}{c}{CIFAR10} & \multicolumn{2}{c}{CIFAR+50} & \multicolumn{2}{c}{Tiny-ImageNet}\\ \cmidrule(r){2-3} \cmidrule(r){4-5} \cmidrule(r){6-7}
         & AUROC & Accuracy & AUROC & Accuracy & AUROC & Accuracy\\ \midrule
         Different & 90.9 & 97.1& 91.0 & 97.3 & 79.7 & 78.5\\
         Random & 91.3 & 97.5& 91.6 & 97.9 & 80.4 & 79.2\\ \bottomrule         
    \end{tabular}
    \label{tab:random}
\end{table}

\begin{table*}[b]
  \centering
  \setlength\tabcolsep{10pt}
  \caption{\revise{Distinguishing in-distribution dataset CIFAR10 from OOD datasets CIFAR100, SVHN, LSUN-Crop, and ImageNet-Crop with different score functions under various metrics.}}
    \begin{tabular}{lllllllll}
    \toprule
    \multirow{2}[1]{*}{Method} & \multicolumn{2}{c}{CIFAR100} & \multicolumn{2}{c}{SVHN} & \multicolumn{2}{c}{LSUN-Crop} & \multicolumn{2}{c}{Imagnet-Crop} \\
    \cmidrule(r){2-3}  \cmidrule(r){4-5} \cmidrule(r){6-7} \cmidrule(r){8-9} & DTACC & AUROC & DTACC & AUROC & DTACC & AUROC & DTACC & AUROC \\
    \midrule
    Plain*~\cite{hendrycks2016a} & 79.8 & 86.3 & 86.4 & 90.6 & 87.9 & 93.7 & 88.6 & 94.5\\
    \quad+ BackMix & {$\text{84.9}_{{\text{5.1}\uparrow}}$} & {$\text{91.3}_{{\text{5.0}\uparrow}}$} & {$\text{88.5}_{{\text{2.1}\uparrow}}$} & {$\text{94.1}_{{\text{3.5}\uparrow}}$} & {$\text{92.2}_{{\text{4.3}\uparrow}}$} & {$\text{96.9}_{{\text{3.2}\uparrow}}$} & {$\text{93.4}_{{\text{4.8}\uparrow}}$} & {$\text{97.8}_{{\text{3.3}\uparrow}}$} \\
    \midrule
    Mahalanobis~\cite{lee2018a} & 79.4 & 86.1 & 95.4& 98.7& 98.9&99.7 & 98.9& 99.9\\
    \quad+ BackMix & {$\text{81.7}_{{\text{2.3}\uparrow}}$} & {$\text{88.1}_{{\text{2.0}\uparrow}}$} & {$\text{95.9}_{{\text{0.5}\uparrow}}$} & {$\text{99.3}_{{\text{0.6}\uparrow}}$} & {$\text{98.9}_{{\text{0.0}-}}$} & {$\text{99.7}_{{\text{0.0}-}}$} & {$\text{99.1}_{{\text{0.2}\uparrow}}$} & {$\text{99.9}_{{\text{0.0}-}}$} \\
    \midrule
    Energy~\cite{liu2020energy} & 80.6 & 86.6 & 86.7 & 90.4 & 88.3 & 94.7 & 88.9 & 95.1\\
    \quad+ BackMix & {$\text{85.6}_{{\text{5.0}\uparrow}}$}  & {$\text{92.2}_{{\text{5.6}\uparrow}}$} & {$\text{89.3}_{{\text{2.6}\uparrow}}$} & {$\text{94.8}_{{\text{3.4}\uparrow}}$} & {$\text{92.2}_{{\text{3.9}\uparrow}}$} & {$\text{97.2}_{{\text{2.5}\uparrow}}$} & {$\text{93.4}_{{\text{4.5}\uparrow}}$} & {$\text{98.1}_{{\text{3.0}\uparrow}}$} \\    
    \midrule
    ODIN~\cite{liang2018enhancing} & 80.4&87.9 & 88.0 & 95.5 & 94.3& 98.5& 96.0 & 99.3\\
    \quad+ BackMix & {$\text{85.0}_{{\text{4.6}\uparrow}}$} & {$\text{92.1}_{{\text{4.2}\uparrow}}$} & {$\text{95.2}_{{\text{7.2}\uparrow}}$} & {$\text{99.0}_{{\text{3.5}\uparrow}}$} & {$\text{96.4}_{{\text{2.1}\uparrow}}$} & {$\text{99.3}_{{\text{0.8}\uparrow}}$} & {$\text{97.7}_{{\text{1.7}\uparrow}}$} & {$\text{99.7}_{{\text{0.4}\uparrow}}$} \\
    \bottomrule
    \end{tabular}
  \label{tab:score}%
\end{table*}%
\textbf{Classifier training.} For each dataset variant, we trained a ResNet18~\cite{he2016deep} with a learning rate of 0.1, batch size of 128, and a cosine annealing learn rate scheduler. We applied a SGD optimizer with Nesterov momentum of 0.9 and weight decay of 5e-4 is adopted for training. And standard data augmentations are used, \ie, first resize the short side of the image to 256 then apply RandomHorizontalFlip and RandomCrop.

\subsection{Experiments for Outlier Exposure}
\label{A_2_2}
\textbf{{Setups.}} In Section~\ref{sec:3.2.2} we followed the experimental setups as well as the training framework in OE~\cite{hendrycks2018deep}. The known outlier dataset, TinyImages, is directly inherited from \cite{hendrycks2018deep}, which is obtained from the 80 Million Tiny Images~\cite{tinyimages80m} with images that appear in CIFAR10 removed. We adopted WideResNet40-4~\cite{zagoruyko2016wide} in these experiments with learn rate 0.1, batch size 128 and a cosine annealing learn rate scheduler. We applied a SGD optimizer with Nesterov momentum of 0.9 and weight decay of 5e-4 was adopted for training.

\section{Supplemental Experiments}
\subsection{The Selection of Available Backgrounds}
\label{A_3_1}
To investigate the impact of background selection within datasets on model performance, we tested the effects of using only images from different known classes (Different) as background regularization and randomly using any image within the batch (Random) as regularization on the CIFAR10 and Tiny-ImageNet datasets under the unknown detection settings.

Results in~\tabref{tab:random} indicate that using only images from different classes as background images for mixing has a relatively small impact on the model performance. The slight performance drop in the Different setting may be due to not mixing target images with the same label background images that are selected randomly within a batch. This also suggests that BackMix establishes the robustness in the selection of backgrounds.

\begin{table}[t]
    \centering
  \setlength\tabcolsep{3.25pt}
    \caption{\revise{The transferability of backbones pretrained using BackMix and other data augmentation methods on multiple visual downstream tasks with different methods.}}
    \begin{tabular}{lp{1.25cm}p{2.25cm}p{1.25cm}p{1.25cm}}
    \toprule
    \multirow{3}{*}{Augmentation} & \multicolumn{2}{c}{Object Detection} & \multicolumn{2}{c}{Image Captioning}\\ \cmidrule(r){2-3} \cmidrule(r){4-5}
    & SSD~\cite{liu2016ssd}\newline (mAP) & Faster-RCNN~\cite{ren2016faster}\newline(mAP) & NIC~\cite{vinyals2015show}\newline (BLEU-1) & NIC~\cite{vinyals2015show}\newline (BLEU-4) \\ \midrule
    Plain* & 76.7 & 75.6 & 61.4 & 22.9 \\
    Mixup~\cite{zhang2017mixup} & {$\text{76.6}_{{\text{0.1}\downarrow}}$} & {$\text{73.9}_{{\text{1.7}\downarrow}}$} & {$\text{61.6}_{{\text{0.2}\uparrow}}$} & {$\text{23.2}_{{\text{0.3}\uparrow}}$} \\
    Cutout~\cite{devries2017improved} & {$\text{76.8}_{{\text{0.1}\uparrow}}$} & {$\text{75.0}_{{\text{0.6}\downarrow}}$} & {$\text{63.0}_{{\text{1.6}\uparrow}}$} & {$\text{24.0}_{{\text{1.1}\uparrow}}$}\\
    Cutmix~\cite{yun2019cutmix} & {$\text{77.6}_{{\text{0.9}\uparrow}}$} & {$\text{76.7}_{{\text{1.1}\uparrow}}$} & {$\text{64.2}_{{\text{2.8}\uparrow}}$} & {$\text{24.9}_{{\text{2.0}\uparrow}}$} \\ \midrule
    BackMix & {$\text{77.9}_{{\text{1.2}\uparrow}}$} & {$\text{77.1}_{{\text{1.5}\uparrow}}$} & {$\text{68.5}_{{\text{7.1}\uparrow}}$} & {$\text{25.6}_{{\text{2.7}\uparrow}}$}  \\
    \bottomrule
    \end{tabular}
    \label{tab:transfer}
\end{table}

\begin{table}[t]
    \centering
  \setlength\tabcolsep{4.5pt}
    \caption{\revise{Comparison of various image captioning metrics for backbones pretrained using BackMix and other data augmentation methods on the COCO dataset.}}
    \begin{tabular}{llllll}
    \toprule
    {Augmentation} & BLEU-2 & BLEU-3 & METEOR & ROUGE-L & CIDEr\\ \midrule
    Plain* &  43.8 & 31.4 & 22.8 & 44.7 & 71.2 \\
    Mixup~\cite{zhang2017mixup} &  {$\text{44.1}_{{\text{0.3}\uparrow}}$} & {$\text{31.6}_{{\text{0.2}\uparrow}}$} & {$\text{22.9}_{{\text{0.1}\uparrow}}$} & {$\text{47.9}_{{\text{3.2}\uparrow}}$} & {$\text{72.2}_{{\text{1.0}\uparrow}}$}   \\
    Cutout~\cite{devries2017improved} & {$\text{45.3}_{{\text{1.5}\uparrow}}$} & {$\text{32.6}_{{\text{1.2}\uparrow}}$}  & {$\text{22.6}_{{\text{0.2}\downarrow}}$} & {$\text{48.2}_{{\text{3.5}\uparrow}}$} & {$\text{74.1}_{{\text{2.9}\uparrow}}$} \\
    Cutmix~\cite{yun2019cutmix} & {$\text{46.3}_{{\text{2.5}\uparrow}}$} & {$\text{33.6}_{{\text{2.2}\uparrow}}$}  & {$\text{23.1}_{{\text{0.3}\uparrow}}$} & {$\text{49.0}_{{\text{4.3}\uparrow}}$} & {$\text{77.6}_{{\text{6.4}\uparrow}}$} \\ \midrule
    BackMix & {$\text{50.6}_{{\text{6.8}\uparrow}}$} & {$\text{36.1}_{{\text{4.7}\uparrow}}$}  & {$\text{23.1}_{{\text{0.3}\uparrow}}$} & {$\text{50.2}_{{\text{5.5}\uparrow}}$} & {$\text{80.6}_{{\text{9.4}\uparrow}}$} \\
    \bottomrule
    \end{tabular}
    \label{tab:caption}
\end{table}

\subsection{\revise{BackMix with Different Score Functions}}
\label{A_3_2}
\revise{As highlighted in~\cite{ming2022impact}, using different score functions may offer alternative perspectives in handling spurious correlations. We extended our analysis by testing the proposed method with the baseline MSP (Plain*)~\cite{hendrycks2016a}, prediction-based methods ODIN~\cite{liang2018enhancing}, Energy~\cite{liu2020energy}, and feature-based method Mahalanobis Distance~\cite{lee2018a}. We used CIFAR10 as the in-distribution (InD) dataset, with CIFAR100, SVHN, LSUN-Crop, and ImageNet-Crop as the out-of-distribution (OOD) datasets. We recorded the DTACC and AUROC values of the baseline model under different score functions, with and without the BackMix.}

\begin{figure}[t]
    \centering
\subfigure[]{  
        \includegraphics[width=0.475\textwidth]{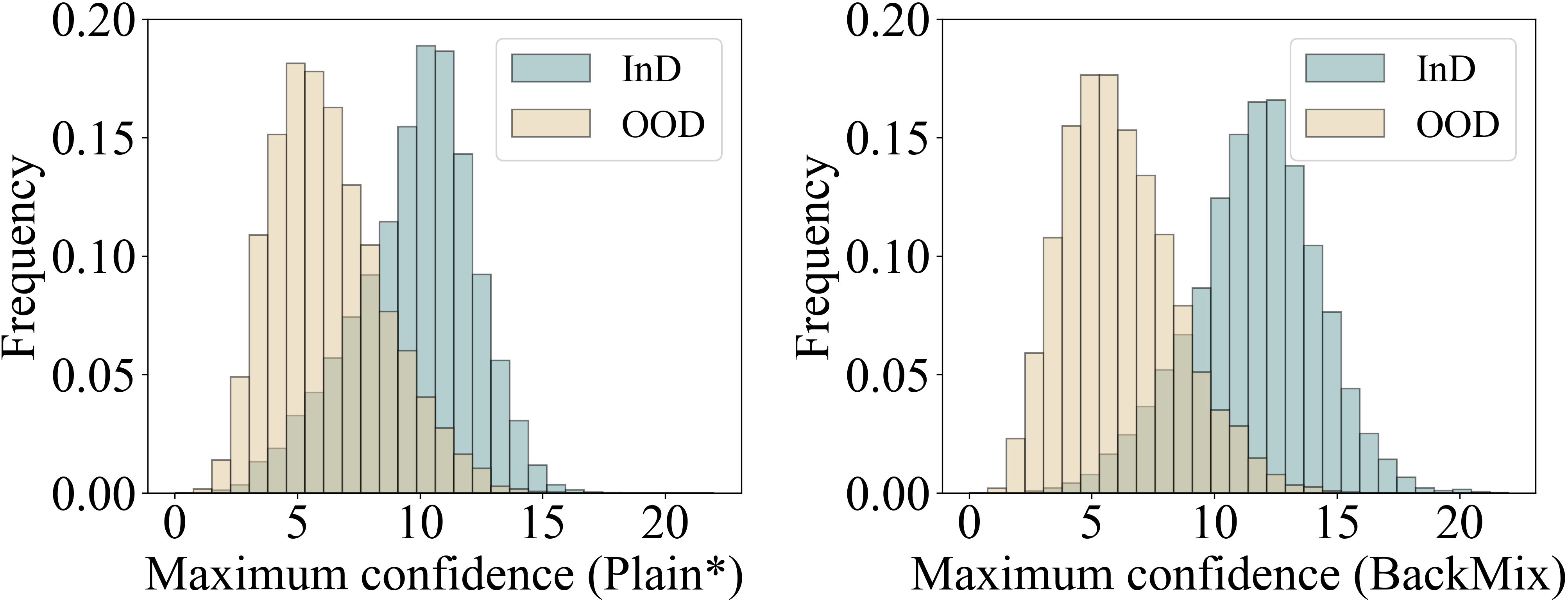}
        }
\subfigure[]{  
        \includegraphics[width=0.475\textwidth]{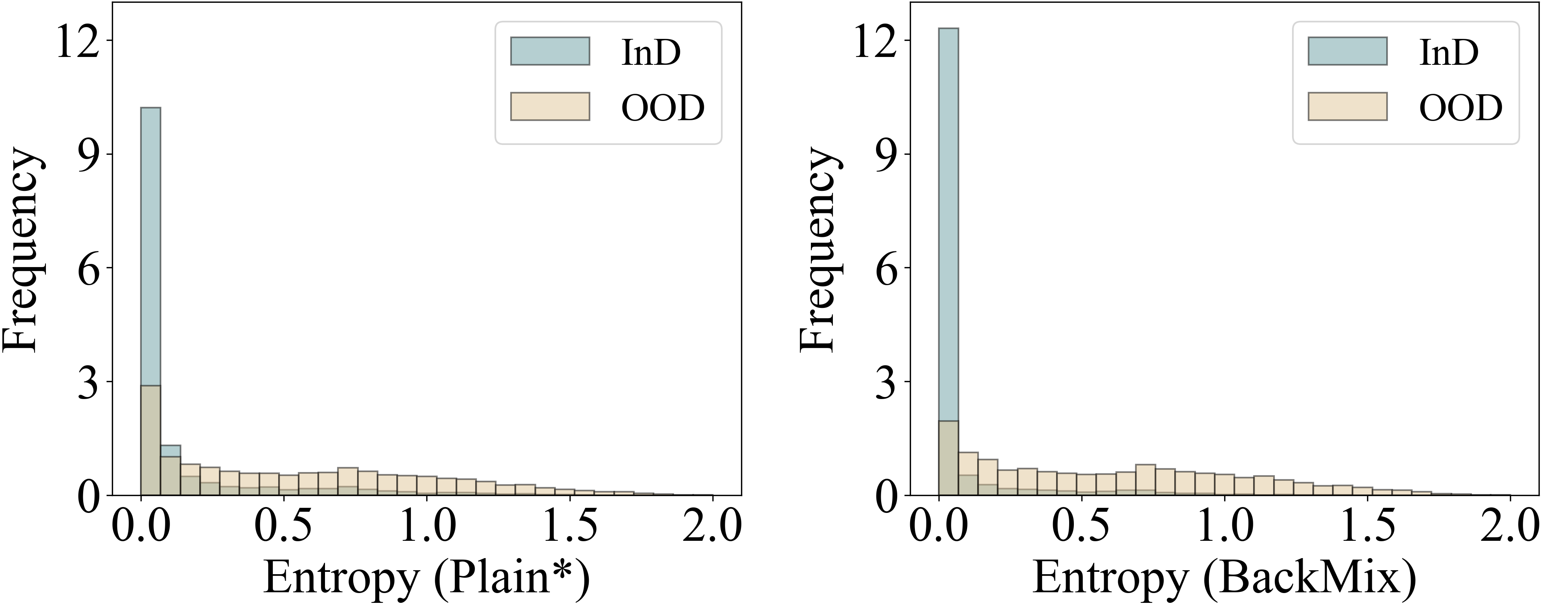}
        }
    \caption{Comparison of the (a) maximum confidence score and the (b) prediction probability entropy output by Plain* and BackMix models on the in-distribution dataset CIFAR10 and the OOD dataset CIFAR100.}
    \label{fig:score_entropy}
\end{figure}

\revise{\textbf{BackMix enables the model to learn the foreground object of known samples and reduces the uncertainty of predictions, further enhancing the effectiveness of prediction-based and feature-based score functions.} Results in~\tabref{tab:score} show that BackMix establishes stable performance improvement with different score functions.~\figref{fig:score_entropy} illustrates the highest confidence of the baseline method (Plain*) and BackMix output, along with the entropy of the prediction probabilities normalized by SoftMax. This indicates that BackMix can reduce the uncertainty of the model’s predictions, and thus improve the model’s discriminative ability for known and unknown samples. Consequently, it can further enhance the performance of prediction-based score functions.}

\revise{Moreover, we observed that the Mahalanobis distance~\cite{lee2018a}, which measures the distance of test samples from known samples in feature space, performs well on far OOD datasets such as SVHN, but diminishes effectiveness in the near OOD dataset CIFAR100, where similar feature distributions limit its ability to separate samples. BackMix can also effectively enhance the Mahalanobis distance method, verifying its ability to help the model learn foreground features.}

\subsection{\revise{The Transferability of BackMix}}
\label{A_3_3}
\revise{Following the Cutmix~\cite{yun2019cutmix} setup, we applied BackMix on ResNet50~\cite{he2016deep} for data augmentation during the ImageNet1K classification pretraining phase and then finetuned the trained model on the downstream visual tasks of object detection and image captioning.}

\revise{According to~\cite{yun2019cutmix}, we finetuned the pretrained models on the object detection using SSD~\cite{liu2016ssd} and Faster-RCNN~\cite{ren2016faster} algorithms. The Pascal VOC 2007 and 2012~\cite{everingham2010pascal} \texttt{trainval} was used as training data and we evaluated the model on the VOC 2007 \texttt{test} data using mean Average Precision (mAP) as the evaluation metric, which measures the average precision at different recall levels across all classes. We conducted image captioning experiments on the COCO dataset~\cite{cocods} using NIC~\cite{vinyals2015show} with the model pretrained with various data augmentation methods. The Bilingual Evaluation Understudy (BLEU) -n is used to evaluate the generated captions by comparing n-gram matches, where `n' represents the length of contiguous word sequences. We presented results in Table~\ref{tab:transfer}, where the results of methods other than ours are taken from~\cite{yun2019cutmix}. Compared to other methods, BackMix delivers greater performance improvements across different algorithms and downstream tasks. }

\revise{We provided more metrics comparison on image captioning task in Table~\ref{tab:caption}. METEOR evaluates the quality of natural language generation tasks. ROUGE-L is a metric used to evaluate the quality of generated text by measuring the longest common subsequence between the generated and reference text. CIDEr evaluates image captions by comparing how much the generated caption matches human-written reference captions. Results demonstrate that BackMix delivers significant performance improvements over the baseline methods across various metrics.}

\begin{figure}[t]
\centering
\includegraphics[width=\linewidth]{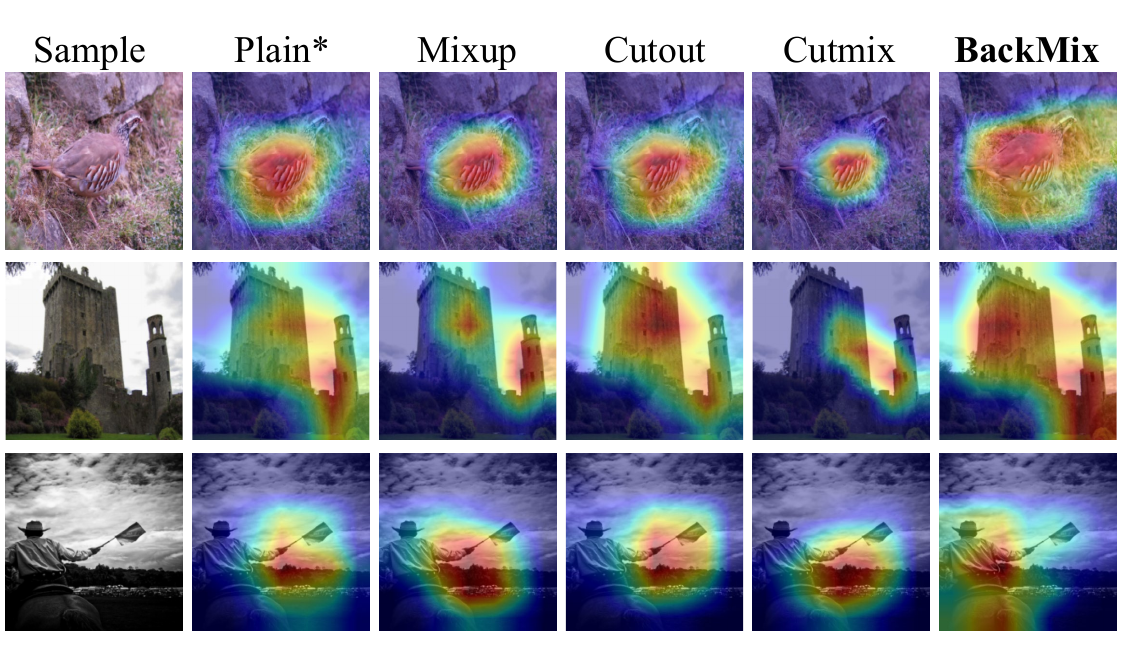}
\caption{\revise{Comparison of Grad-CAM results on the ImageNet1K dataset using different data augmentation methods.}}
\vspace{-5mm}
\label{fig:aug_cam}
\end{figure}

\begin{table*}[t]
  \centering
  \setlength\tabcolsep{10pt}
  \caption{\revise{The closed-set and open set performance comparison of applying BackMix to multiple finetuning methods on pretrained models for the out-of-distribution detection task. CIFAR10 was used as the in-distribution dataset, while CIFAR100 and SVHN were used as out-of-distribution datasets.}}
    \begin{tabular}{lllllllll}
    \toprule
    \multirow{2}[1]{*}{Setting} & \multirow{2}[1]{*}{Method} & & \multicolumn{3}{c}{CIFAR100} & \multicolumn{3}{c}{SVHN} \\
    \cmidrule(r){4-6}  \cmidrule(r){7-9} & & Accuracy & AUROC & OSCR & Macro-F1 & AUROC & OSCR & Macro-F1 \\
    \midrule
    \multirow{4}{*}{1-shot} 
    & CoOp & 89.8&91.6 &85.0&74.4& 98.8 & 89.5 & 86.3\\
    & \quad+ BackMix & {$\text{90.7}_{{\text{0.9}\uparrow}}$} &  {$\text{92.1}_{{\text{0.5}\uparrow}}$} &  {$\text{85.8}_{{\text{0.8}\uparrow}}$} & {$\text{75.4}_{{\text{1.0}\uparrow}}$} &  {$\text{99.0}_{{\text{0.2}\uparrow}}$} &  {$\text{90.3}_{{\text{0.8}\uparrow}}$} &
    {$\text{87.4}_{{\text{1.1}\uparrow}}$} \\
    \cmidrule(r){2-9}
    & LoCoOp & 89.6 & 91.2 & 84.6 &  74.2 &  98.5 &  89.0 & 85.1\\
    & \quad+ BackMix & {$\text{90.7}_{{\text{1.1}\uparrow}}$} &  {$\text{91.6}_{{\text{0.4}\uparrow}}$} &  {$\text{85.2}_{{\text{0.6}\uparrow}}$} &  {$\text{76.0}_{{\text{1.8}\uparrow}}$} & 
    {$\text{99.0}_{{\text{0.5}\uparrow}}$} & 
    {$\text{90.1}_{{\text{1.1}\uparrow}}$} & 
    {$\text{88.0}_{{\text{2.9}\uparrow}}$} \\
    \midrule
    \multirow{4}{*}{4-shot} 
    & CoOp & 90.6 & 91.5 & 85.3 & 74.4 & 99.0 & 90.1 & 87.3\\
    & \quad+ BackMix &  {$\text{91.3}_{{\text{0.7}\uparrow}}$} &  {$\text{92.1}_{{\text{0.6}\uparrow}}$} &  {$\text{86.3}_{{\text{1.0}\uparrow}}$}&  {$\text{76.0}_{{\text{1.6}\uparrow}}$} &  {$\text{99.1}_{{\text{0.1}\uparrow}}$}&  {$\text{90.8}_{{\text{0.7}\uparrow}}$}& 
     {$\text{88.3}_{{\text{1.0}\uparrow}}$}\\
    \cmidrule(r){2-9}
    & LoCoOp & 89.8 & 91.4 & 84.6 & 73.8 & 98.6 & 89.2 & 84.8\\
    & \quad+ BackMix & {$\text{90.9}_{{\text{1.1}\uparrow}}$}&  {$\text{91.9}_{{\text{0.5}\uparrow}}$} & {$\text{85.7}_{{\text{1.1}\uparrow}}$}&  {$\text{76.1}_{{\text{2.3}\uparrow}}$} & {$\text{99.0}_{{\text{0.4}\uparrow}}$}&   {$\text{90.4}_{{\text{1.2}\uparrow}}$}&   {$\text{87.9}_{{\text{3.1}\uparrow}}$}\\
    \midrule
    \multirow{4}{*}{16-shot} & CoOp & 91.2 & 91.1 & 85.6 & 74.5 & 98.6 & 90.7 & 85.8\\
    & \quad+ BackMix &  {$\text{91.7}_{{\text{0.5}\uparrow}}$} &  {$\text{91.6}_{{\text{0.5}\uparrow}}$} &  {$\text{86.3}_{{\text{0.7}\uparrow}}$} & {$\text{75.8}_{{\text{1.3}\uparrow}}$} &  {$\text{99.0}_{{\text{0.4}\uparrow}}$} &  {$\text{91.3}_{{\text{0.6}\uparrow}}$}&  {$\text{87.7}_{{\text{1.9}\uparrow}}$}\\
    \cmidrule(r){2-9}
    & LoCoOp & 91.4 & 90.4 & 85.5 & 71.9 & 93.7 & 87.6 &79.8\\
    & \quad+ BackMix & {$\text{91.7}_{{\text{0.3}\uparrow}}$} & {$\text{91.0}_{{\text{0.6}\uparrow}}$} & {$\text{86.3}_{{\text{0.8}\uparrow}}$} & {$\text{73.8}_{{\text{1.9}\uparrow}}$} & {$\text{96.1}_{{\text{2.4}\uparrow}}$} & {$\text{89.7}_{{\text{2.1}\uparrow}}$} & {$\text{82.5}_{{\text{2.7}\uparrow}}$} \\
    \midrule
    \multirow{6}{*}{Full-data}& CoOp & 93.2 & 91.1 & 87.2 & 73.5 & 98.8 & 92.6 & 88.3\\
    & \quad+ BackMix & {$\text{94.0}_{{\text{0.8}\uparrow}}$} &{$\text{92.7}_{{\text{1.6}\uparrow}}$} & {$\text{89.0}_{{\text{1.8}\uparrow}}$} & {$\text{77.7}_{{\text{4.2}\uparrow}}$} & {$\text{98.7}_{{\text{0.1}\downarrow}}$} & {$\text{93.4}_{{\text{0.8}\uparrow}}$} & {$\text{88.7}_{{\text{0.4}\uparrow}}$}\\ \cmidrule(r){2-9}
    & LoCoOp & 94.3 & 93.1 & 89.6 & 77.8 & 95.3 & 91.2 & 84.1\\ 
    & \quad+ BackMix & {$\text{94.6}_{{\text{0.3}\uparrow}}$} & {$\text{93.3}_{{\text{0.2}\uparrow}}$} & {$\text{90.0}_{{\text{0.4}\uparrow}}$} & {$\text{78.7}_{{\text{0.9}\uparrow}}$} & {$\text{97.3}_{{\text{2.0}\uparrow}}$} & {$\text{93.3}_{{\text{2.1}\uparrow}}$} & {$\text{87.0}_{{\text{2.9}\uparrow}}$}\\\cmidrule(r){2-9}
    & VPT & 96.2 & 95.0 & 92.4 & 82.5 & 97.5 & 94.4 & 84.6\\
    & \quad+ BackMix & {$\text{96.5}_{{\text{0.3}\uparrow}}$} & {$\text{95.5}_{{\text{0.5}\uparrow}}$} & {$\text{93.0}_{{\text{0.6}\uparrow}}$} & {$\text{83.8}_{{\text{1.3}\uparrow}}$} & {$\text{98.3}_{{\text{0.8}\uparrow}}$} & {$\text{95.0}_{{\text{0.6}\uparrow}}$} & {$\text{85.6}_{{\text{1.0}\uparrow}}$}\\    
    \bottomrule
    \end{tabular}
\vspace{-3mm}
  \label{tab:a_ptm}%
\end{table*}%

\revise{Fig.~\ref{fig:aug_cam} shows the Grad-CAM~\cite{selvaraju2017grad} of models trained using baseline strategy (Plain*) and different data augmentation techniques on test images from the ImageNet1K dataset. Compared to other methods, BackMix exhibits more comprehensive attention and can accurately focus on the main body of the corresponding label class in scenes with multiple objects. As mentioned in representation learning methods~\cite{caron2021emerging,oquab2024dinov2}, a backbone with better feature extraction capabilities can perform better on downstream tasks. Therefore, the more noticeable performance gains of BackMix in various visual downstream tasks may be attributed to its enhancement of the model’s representation capabilities.}

\begin{table*}[t]
    \centering
    \setlength\tabcolsep{10pt}
    \caption{\revise{Comparison of the baseline and different score functions combined with BackMix on the WaterBirds dataset with varying degrees of spurious fore-background correlation (Corr.).}}
    \begin{tabular}{lcllllllll}
    \toprule
    \multirow{2}{*}{Method} & \multirow{2}{*}{Corr.} & \multicolumn{2}{c}{Spurious OOD}  & \multicolumn{2}{c}{SVHN} & \multicolumn{2}{c}{iSUN} & \multicolumn{2}{c}{LSUN}\\ 
    \cmidrule(r){3-4} \cmidrule(r){5-6} \cmidrule(r){7-8} \cmidrule(r){9-10}
    & & FPR95$\downarrow$& AUROC$\uparrow$& FPR95$\downarrow$& AUROC$\uparrow$& FPR95$\downarrow$& AUROC$\uparrow$& FPR95$\downarrow$& AUROC$\uparrow$\\ \midrule
    Plain* & \multirow{8}{*}{$r$ = 0.5} & 55.7 & 90.2 & 12.7 & 96.6 & 19.0 & 95.2 &16.1 & 94.3  \\
    \quad + BackMix & & $\text{55.3}_{\text{0.4}\downarrow}$ &$\text{90.8}_{\text{0.6}\uparrow}$ &$\text{12.3}_{\text{0.4}\downarrow}$ &$\text{97.1}_{\text{0.5}\uparrow}$ &$\text{18.3}_{\text{0.7}\downarrow}$ &$\text{95.5}_{\text{0.3}\uparrow}$ &$\text{15.8}_{\text{0.3}\downarrow}$ &$\text{94.8}_{\text{0.5}\uparrow}$ \\ \cmidrule(r){1-1} \cmidrule(r){3-10} 
    Mahalanobis & & 56.1 & 86.3 & 0.5 & 100.0 & 0.6 & 99.9 & 0.6 & 100.0 \\
    \quad + BackMix & & $\text{55.7}_{\text{0.4}\downarrow} $ &$\text{87.0}_{\text{0.7}\uparrow} $ &$\text{0.2}_{\text{0.3}\downarrow} $ &$\text{100.0}_{\text{0.0}-} $ &$\text{0.2}_{\text{0.4}\downarrow} $ &$\text{100.0}_{\text{0.1}\uparrow} $ &$\text{0.3}_{\text{0.3}\downarrow} $ &$\text{100.0}_{\text{0.0}-}$\\ \cmidrule(r){1-1} \cmidrule(r){3-10} 
    Energy & & 54.9 & 90.5 & 6.5 & 98.9 & 10.2 & 97.5 & 12.6 & 97.2\\
    \quad + BackMix & & $\text{54.6}_{\text{0.3}\downarrow} $ &$\text{91.0}_{\text{0.5}\uparrow} $ &$\text{6.4}_{\text{0.1}\downarrow} $ &$\text{99.0}_{\text{0.1}\uparrow} $ &$\text{10.0}_{\text{0.2}\downarrow} $ &$\text{97.5}_{\text{0.0}-} $ &$\text{12.3}_{\text{0.3}\downarrow} $ &$\text{97.4}_{\text{0.2}\uparrow} $ \\ \cmidrule(r){1-1} \cmidrule(r){3-10} 
    ODIN & & 54.3 & 90.5 & 6.1 & 99.1 & 10.1 & 97.5 & 12.7 & 96.9\\
    \quad + BackMix & & $\text{54.3}_{\text{0.0}-} $ &$\text{90.7}_{\text{0.2}\uparrow} $ &$\text{6.0}_{\text{0.1}\downarrow} $ &$\text{99.3}_{\text{0.2}\uparrow} $ &$\text{10.1}_{\text{0.0}-} $ &$\text{98.0}_{\text{0.5}\uparrow} $ &$\text{12.4}_{\text{0.3}\downarrow} $ &$\text{97.3}_{\text{0.4}\uparrow} $\\ \midrule 
    Plain* & \multirow{8}{*}{$r$ = 0.7} & 73.0 & 81.3 & 37.4 & 94.8 & 46.7 & 91.6 & 46.8 & 91.6 \\
    \quad + BackMix & & $\text{71.5}_{\text{1.5}\downarrow} $ &$\text{85.1}_{\text{3.8}\uparrow} $ &$\text{35.8}_{\text{1.6}\downarrow} $ &$\text{95.8}_{\text{1.0}\uparrow} $ &$\text{43.3}_{\text{3.4}\downarrow} $ &$\text{92.7}_{\text{1.1}\uparrow} $ &$\text{44.0}_{\text{2.8}\downarrow} $ &$\text{93.1}_{\text{1.5}\uparrow} $ \\ \cmidrule(r){1-1} \cmidrule(r){3-10} 
    Mahalanobis & & 71.8 & 82.9 & 0.5 & 99.9 & 0.9 & 99.8 & 1.1 & 99.8 \\
    \quad + BackMix & & $\text{71.3}_{\text{0.5}\downarrow} $ &$\text{83.6}_{\text{0.7}\uparrow} $ &$\text{0.2}_{\text{0.3}\downarrow} $ &$\text{100.0}_{\text{0.1}\uparrow} $ &$\text{0.6}_{\text{0.3}\downarrow} $ &$\text{99.9}_{\text{0.1}\uparrow} $ &$\text{0.7}_{\text{0.4}\downarrow} $ &$\text{100.0}_{\text{0.2}\uparrow} $ \\ \cmidrule(r){1-1} \cmidrule(r){3-10} 
    Energy & & 72.5 & 84.6 & 36.9 & 95.4 & 41.9 & 91.9 & 40.4 & 92.4 \\
    \quad + BackMix & &$\text{71.3}_{\text{1.2}\downarrow} $ &$\text{86.1}_{\text{1.5}\uparrow} $ &$\text{35.3}_{\text{1.6}\downarrow} $ &$\text{96.1}_{\text{0.7}\uparrow} $ &$\text{39.5}_{\text{2.4}\downarrow} $ &$\text{93.3}_{\text{1.4}\uparrow} $ &$\text{38.6}_{\text{1.8}\downarrow} $ &$\text{93.7}_{\text{1.3}\uparrow} $\\ \cmidrule(r){1-1} \cmidrule(r){3-10} 
    ODIN & & 73.6 & 81.3 & 36.4 & 95.5 & 42.1 & 92.0 & 40.9 & 92.3 \\
    \quad + BackMix & & $\text{71.9}_{\text{1.7}\downarrow} $ &$\text{83.8}_{\text{2.5}\uparrow} $ &$\text{35.3}_{\text{1.1}\downarrow} $ &$\text{96.2}_{\text{0.7}\uparrow} $ &$\text{39.8}_{\text{2.3}\downarrow} $ &$\text{93.1}_{\text{1.1}\uparrow} $ &$\text{39.1}_{\text{1.8}\downarrow} $ &$\text{93.5}_{\text{1.2}\uparrow} $ \\ \midrule 
    Plain* & \multirow{8}{*}{$r$ = 0.9} & 86.1 & 74.4 & 44.7 & 92.3 & 51.4 & 89.6 & 49.5 & 90.1 \\
    \quad + BackMix & & $\text{81.5}_{\text{4.6}\downarrow} $ &$\text{80.3}_{\text{5.9}\uparrow} $ &$\text{42.6}_{\text{2.1}\downarrow} $ &$\text{93.2}_{\text{0.9}\uparrow} $ &$\text{49.5}_{\text{1.9}\downarrow} $ &$\text{91.0}_{\text{1.4}\uparrow} $ &$\text{46.2}_{\text{3.2}\downarrow} $ &$\text{91.9}_{\text{1.8}\uparrow} $ \\ \cmidrule(r){1-1} \cmidrule(r){3-10} 
    Mahalanobis & & 79.5 & 76.3 & 0.9 & 99.8 & 1.1 & 99.6 & 1.9 & 99.5\\
    \quad + BackMix & & $\text{76.2}_{\text{3.3}\downarrow} $ &$\text{80.5}_{\text{4.2}\uparrow} $ &$\text{0.4}_{\text{0.5}\downarrow} $ &$\text{100.0}_{\text{0.2}\uparrow} $ &$\text{0.7}_{\text{0.4}\downarrow} $ &$\text{99.9}_{\text{0.3}\uparrow} $ &$\text{0.8}_{\text{1.1}\downarrow} $ &$\text{99.9}_{\text{0.4}\uparrow}$ \\ \cmidrule(r){1-1} \cmidrule(r){3-10} 
    Energy & & 84.5 & 75.2 & 44.3 & 92.5 & 50.9 & 90.0 & 49.7 & 89.8\\
    \quad + BackMix & & $\text{81.7}_{\text{2.8}\downarrow} $ &$\text{80.6}_{\text{5.4}\uparrow} $ &$\text{42.4}_{\text{1.9}\downarrow} $ &$\text{93.8}_{\text{1.3}\uparrow} $ &$\text{49.4}_{\text{1.5}\downarrow} $ &$\text{91.1}_{\text{1.1}\uparrow} $ &$\text{46.6}_{\text{3.1}\downarrow} $ &$\text{91.5}_{\text{1.7}\uparrow} $\\ \cmidrule(r){1-1} \cmidrule(r){3-10} 
    ODIN & & 84.8 & 74.8 & 44.6 & 92.6 & 50.8 & 90.0 & 49.2 & 90.4\\
    \quad + BackMix & & $\text{81.6}_{\text{3.2}\downarrow} $ &$\text{79.9}_{\text{5.1}\uparrow} $ &$\text{42.5}_{\text{2.1}\downarrow} $ &$\text{94.1}_{\text{1.5}\uparrow} $ &$\text{49.8}_{\text{1.0}\downarrow} $ &$\text{91.2}_{\text{1.2}\uparrow} $ &$\text{46.0}_{\text{3.2}\downarrow} $ &$\text{91.6}_{\text{1.2}\uparrow} $  \\ 
    \bottomrule
    \end{tabular}
    \label{tab:spurious}
\end{table*}

\subsection{\revise{BackMix on Large-scale Pretrained Models}}
\label{A_3_4}
\revise{To validate the effectiveness of our approach on large-scale pretrained models, we conducted experiments on the pure ViT-based finetuning method VPT~\cite{jia2022vpt}, as well as the CLIP-based finetuning methods CoOp~\cite{zhou2022coop} and LoCoOp~\cite{miyai2023locoop}. VPT introduces prompts within visual models to adapt them without altering backbone weights. CoOp finetunes CLIP by using learnable prompts and LoCoOp further uses local features for OOD regularization.}

\revise{We evaluated the CoOp and LoCoOp across 1-shot, 4-shot, 16-shot and full-data settings. Considering that VPT requires training a classification head, we evaluated its performance in the full-data setting. CIFAR10 served as the InD dataset, and CIFAR100 and SVHN served as OOD datasets. We adopted the ViT-B/16 model trained by~\cite{radford2021clip} as the backbone for all methods and input both original and processed data for BackMix.}

\revise{\textbf{BackMix consistently improves the performance of pretrained model across data scales, proving its strength in the finetuning stage.} Results in~\tabref{tab:a_ptm} indicate that BackMix can further enhance the pretrained model's closed-set classification and unknown detection capabilities across different scales of training data, even in the scenario where only one training sample from each class is available. Benefiting from the powerful feature extraction capabilities of the pretrained models, BackMix can more accurately distinguish between the background and foreground of images, thereby achieving superior performance.}

\subsection{\revise{BackMix on Mitigating Spurious Correlations}}
\label{A_3_5}
\revise{Following the settings in~\cite{ming2022impact}, we combined various OOD detection methods with BackMix and tested these methods on the Waterbirds~\cite{sagawa2019distributionally} dataset with different spurious correlation $r$ to evaluate the performance of BackMix on mitigating impacts of spurious correlation. As proposed in~\cite{ming2022impact}, the correlation $r$ is defined as:
\begin{align}
r&=P(e=\text{water} | y = \text{waterbirds})\notag \\&=P(e=\text{land} |y = \text{landbirds}),
\end{align}
where $e$ denotes the label of environment and $y$ denotes the label of foreground object. For this dataset, $r$=0.5 indicates that classes appear uniformly across different backgrounds, reflecting a low level of spurious correlation, while $r$=0.9 means that classes almost exclusively appear in strongly correlated backgrounds, indicating a high level of spurious correlation. We used a subset of images of land and water from the Places dataset as the `Spurious OOD' dataset, while using SVHN, iSUN, and LSUN as non-spurious datasets. FPR95 and AUROC are adopted as evaluation metrics. FPR95 measures the false positive rate when 95\% of the known samples are correctly accepted.}

\revise{\textbf{BackMix can mitigate impacts of spurious correlations.} When $r$=0.7 and $r$=0.9, spurious correlations exist in the data. Results in Table~\ref{tab:spurious} indicate that BackMix reduces the proportion of misclassified unknown samples with spurious correlations and improves the overall performance of the model. Particularly for Spurious OOD, BackMix significantly enhances the model's detection capability. This indicates that the fore-background decoupling of BackMix can effectively mitigate the impacts of priors in the dataset. Moreover, BackMix can be integrated with various OOD detection methods to achieve better performance.}

\revise{\textbf{Mitigating impacts of spurious correlations between foreground and background can improve the model's performance on OSR and OOD detection tasks.} When $r$=0.5, the model performs significantly better than when $r$=0.9. By removing fore-background correlations, the model is able to focus on the primary classification subject better, thus avoiding misclassification caused by unknown data with spurious correlations. This conclusion also validates the effectiveness of BackMix in removing fore-background correlations in the training set for OSR and OOD detection tasks.}

\subsection{More Visualizations}
\label{A_3_6}
\begin{figure*}[t]
        \includegraphics[width=\textwidth]{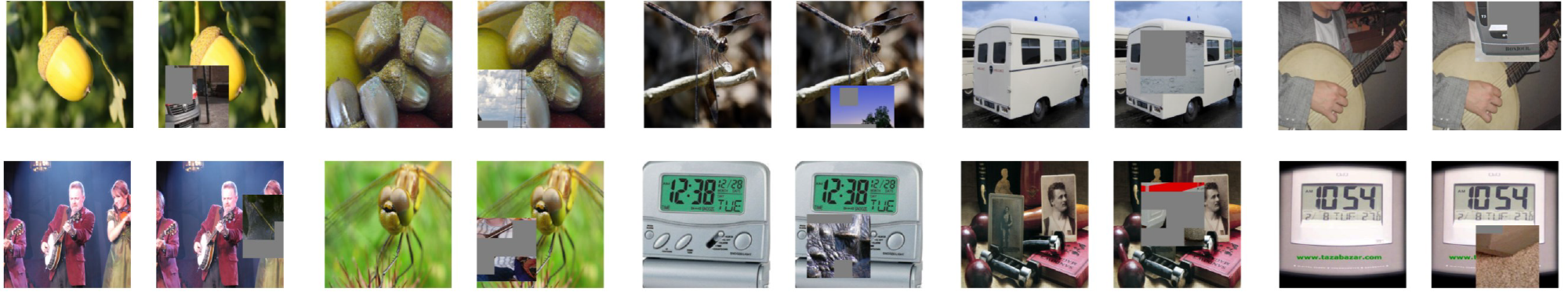}
    \caption{Samples processed with BackMix on the ImageNet30 dataset.}
    \label{fig:vis_suppl}
\end{figure*}

We present more images processed with BackMix in~\figref{fig:vis_suppl}. With the progressively optimized foreground segmentation, the model effectively masks foreground objects in the BI, successfully avoiding the introduction of multiple object classes in a single image after being processed by BackMix.

\begin{table*}[t]
    \caption{Comparison of the key characteristics of representative studies in spurious correlations and BackMix.}
\begin{tabular}{p{2.25cm}p{5cm}p{4.5cm}p{4.5cm}}
    \toprule
    \multirow{2}{*}{Study}& Impacts of Fore-Background Correlation\newline on OSR / OOD Detection & \multirow{2}{*}{Address Fore-Background Correlations} & \multirow{2}{*}{\quad\quad\quad\quad Applicability}\\ \midrule
    Spurious OOD\newline~\cite{ming2022impact} & 1. Analyze impacts on elaborately constructed datasets\newline
2. Feature-based score function mitigates negative impacts &  \multirow{2}{*}{\quad\quad\quad\quad\quad\quad\ding{55}} & \multirow{2}{*}{Limited experimental scenarios}\\ \midrule
Existing Methods\newline~\cite{creager2021environment,yao2022improving,asgari2022masktune,liu2023avoiding} & \multirow{2}{*}{\quad\quad\quad\quad\quad\quad\ding{55}} & \multirow{2}{*}{Focus on foreground feature mainly} & Tasks with spurious correlation priors using additional annotations  \\ \midrule
\multirow{3}{*}{\textbf{Ours}} & 1. Analyze impacts on common datasets\newline 2. Analyze impacts of multiple \newline fore-background processing cases\newline 3. Analyze background regularization & 1. Remove fore-background priors with no additional information\newline 2. Use foreground to predict\newline 3. Use background as available outliers & \multirow{3}{*}{\quad\quad\quad\quad\quad\quad\ding{51}}\\
    \bottomrule
\end{tabular}
    \label{tab:mth_list}
\end{table*}

\section{\revise{Further Discussion}}
\subsection{\revise{Connection between BackMix and Existing Methods}}
\label{A_4_1}
\subsubsection{\revise{Connections to OSR methods}}
\label{A_4_1_1}
\revise{Most existing OSR methods jointly model the foregrounds and backgrounds, thereby suffering from failing to identify the test samples that are partially known, \ie, varying foregrounds or backgrounds. OE methods select an auxiliary dataset that serves as a regularizer to alleviate such a problem, but the dataset requires careful design, which is rarely feasible in real-world applications. Theoretically, we find that the proposed BackMix that removes fore-background priors and uses a GAP regularizer works similarly to OE but does not require any auxiliary data.}

\subsubsection{\revise{Connections to spurious correlation mitigation methods}}
\label{A_4_1_2}
\revise{To better illustrate the contributions of BackMix relative to existing methods, we compare key characteristics of the work by Ming~\etal~\cite{ming2022impact} (Spurious OOD), some representative studies in spurious correlations~\cite{creager2021environment,yao2022improving,asgari2022masktune,liu2023avoiding} and BackMix in Table~\ref{tab:mth_list}. It evaluates whether each study explores the impacts of fore-background correlations on OSR and OOD detection tasks, the way of addressing fore-background correlations, and the applicability to general scenarios. Unlike existing works, our study deeply explore impacts of fore-background priors and the impacts of different types of foregrounds and backgrounds in OSR tasks, respectively. More importantly, we proposed a new method BackMix that uses backgrounds as outlier regularizations with no additional information. We proved that the simple BackMix is equivalent to the OE~\cite{hendrycks2018deep} method which elaborately uses auxiliary outliers.}

\revise{\textbf{Impacts of fore-background correlations on OSR / OOD detection.} Ming~\etal~\cite{ming2022impact} (Row 1) conducted experiments on elaborately constructed datasets with spurious correlations and found that fore-background correlations negatively impact model performance. They experimentally concluded that feature-based score function~\cite{lee2018a} in OOD detection can mitigate negative impacts. In contrast, we tested models directly on standard datasets and analyzed how fore-background correlations affect model performance in general cases. Furthermore, we explored ways to enhance model performance while mitigating fore-background correlations, including cases where there are multiple foregrounds and no background in the image. Based on theoretical and experimental analysis, we proposed a strategy using backgrounds as available outliers for regularization.}

\revise{\textbf{Address fore-background correlations.} Existing methods~\cite{creager2021environment,yao2022improving,asgari2022masktune,liu2023avoiding} (Row 2) mitigate impacts of fore-background correlations on the performance by enabling the model to learn foreground more accurately and comprehensively. However, most of these methods do not directly address the negative impacts of fore-background correlations on OSR and OOD detection tasks. Based on our experiments, we found that using background for regularization can improve the performance of model in OSR and OOD detection tasks by treating backgrounds as available outliers and mitigating the negative impacts of fore-background correlations.}

\revise{\textbf{Applicability.} Ming~\etal~\cite{ming2022impact} found that Mahalanobis method~\cite{lee2018a} is relatively effective on data with spurious correlations and far OOD data. Results in Table~\ref{tab:score} indicate that Mahalanobis method~\cite{lee2018a} still suffers due to the inherent prior correlations between foreground and background features, leading to poor performance in handling complex problems such as near OOD data (CIFAR100). Other methods on addressing spurious correlations require additional information~\cite{creager2021environment,liu2023avoiding,yao2022improving}, model components~\cite{liu2023avoiding}, and training objectives~\cite{creager2021environment,liu2023avoiding,asgari2022masktune}. Compared with the aforementioned methods, BackMix can serve as a general data augmentation method, seamlessly integrating with existing methods to improve OSR and OOD detection performance, thus offering greater applicability.}

Thus, BackMix provides a simpler and more effective method, with significant advantages in its easy deployment and specific alignment with OSR, enhancing robustness against both fore-background correlations and OOD samples.

\begin{table}[t]
    \centering
    \setlength\tabcolsep{1.5pt}
    \caption{\revise{Information of 10 classes with strong fore-background correlations selected from the ImageNet1K.}}
    \begin{tabular}{llc}
    \toprule
        Foreground & Background & Images \\ \midrule
        Hare & Grassland & \begin{minipage}{0.26\textwidth}
                    \includegraphics[height=0.8cm]{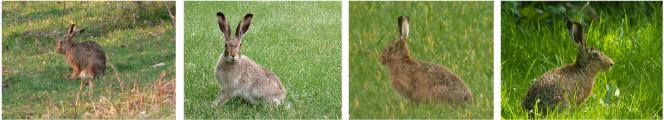}
                  \end{minipage}\\ \midrule
        Hot pot & Cooker& \begin{minipage}{0.26\textwidth}
                    \includegraphics[height=0.8cm]{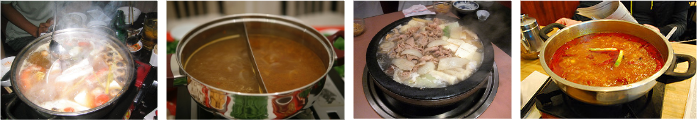}
                  \end{minipage}\\ \midrule
        Taxicab & City road& \begin{minipage}{0.26\textwidth}
                    \includegraphics[height=0.8cm]{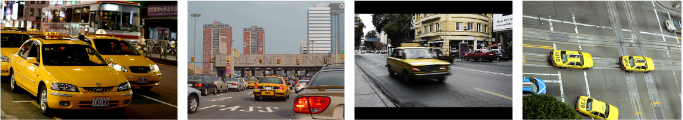}
                  \end{minipage}\\ \midrule
        Cardoon & Leaves& \begin{minipage}{0.26\textwidth}
                    \includegraphics[height=0.8cm]{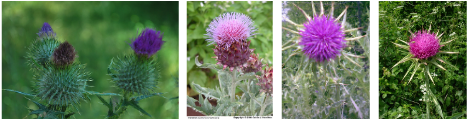}
                  \end{minipage}\\ \midrule
        Dog sled & Snow& \begin{minipage}{0.26\textwidth}
                    \includegraphics[height=0.8cm]{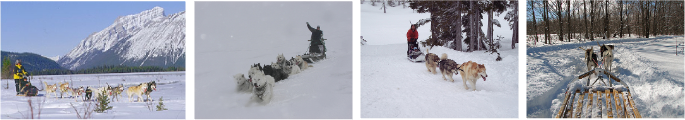}
                  \end{minipage}\\ \midrule
        Parachute & Sky & \begin{minipage}{0.26\textwidth}
                    \includegraphics[height=0.8cm]{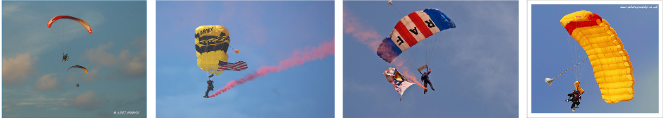}
                  \end{minipage}\\ \midrule
        Container ship & Sea surface& \begin{minipage}{0.26\textwidth}
                    \includegraphics[height=0.79cm]{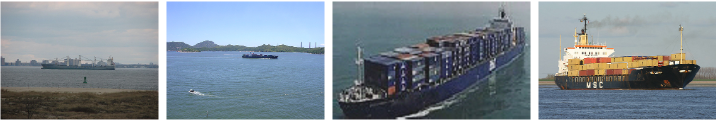}
                  \end{minipage}\\ \midrule 
        Great white shark & Ocean& \begin{minipage}{0.26\textwidth}
                    \includegraphics[height=0.8cm]{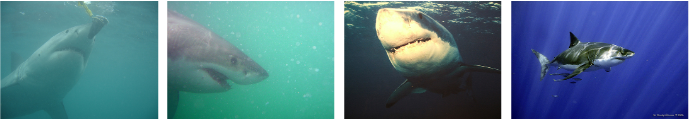}
                  \end{minipage}\\ \midrule
        Entertainment center & Indoor& \begin{minipage}{0.26\textwidth}
                    \includegraphics[height=0.8cm]{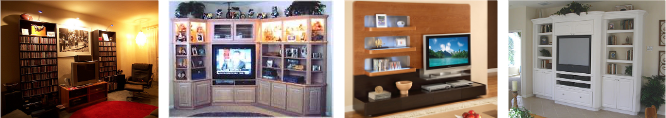}
                  \end{minipage}\\ \midrule
        Lizard & Sand & \begin{minipage}{0.26\textwidth}
                    \includegraphics[height=0.8cm]{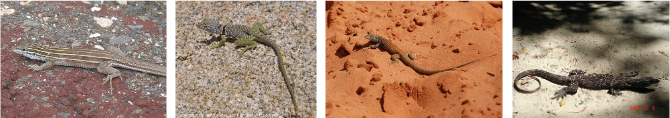}
                  \end{minipage}\\
        \bottomrule
    \end{tabular}
    \label{tab:prior_data}
\end{table}

\begin{table}[t]
    \centering
    \caption{\revise{Comparison of the open set recognition performance of the model on the constructed strong fore-background correlation dataset.}}
    \setlength\tabcolsep{2.25pt}
    \begin{tabular}{lllllll}
    \toprule
       \multirow{2}{*}{Method}  & \multicolumn{2}{c}{4-shot} & \multicolumn{2}{c}{16-shot} & \multicolumn{2}{c}{Full-data} \\
       \cmidrule(r){2-3} \cmidrule(r){4-5} \cmidrule(r){6-7} 
       & Accuracy & AUROC & Accuracy & AUROC & Accuracy & AUROC \\ \midrule
       Plain*  & 53.2 & 64.6 & 67.2 & 65.9 & 96.8& 90.0\\
       \quad + BackMix & $\text{53.2}_\text{0.0}-$ & $\text{65.2}_{\text{0.6}\uparrow}$ & $\text{67.6}_{\text{0.4}\uparrow}$ & $\text{66.8}_{\text{0.9}\uparrow}$ & $\text{98.0}_{\text{1.2}\uparrow}$ & $\text{91.6}_{\text{1.6}\uparrow}$\\
    \bottomrule
    \end{tabular}
    \label{tab:prior_res}
\end{table}

\subsection{\revise{Limitations}}
\label{A_4_2}
\revise{As the BackMix uses CAM to determine the background regions and then applies background regularization to mitigate the influence of fore-background priors in the training set, the performance improvement of BackMix is limited when each foreground only appears in specific background without overlapping, and the background regions extracted by CAM are inaccurate in few-shot learning from scratch.}

\revise{\textbf{Each foreground only appears in a specific background without overlapping.} When each class in a dataset appears only in a fixed background, the performance improvement from BackMix is limited. In such cases, the fore-background priors also apply during test.}

\revise{To simulate this rare scenario, we selected 10 classes from the ImageNet dataset, each with a fixed, non-overlapping background as shown in Table~\ref{tab:prior_data}. Five classes were designated as known, and the remaining five as unknown during test. Table~\ref{tab:prior_res} presents the closed-set classification accuracy and AUROC performance under few-shot and full data settings.}

\revise{With strong fore-background correlations and no overlap, backgrounds can effectively aid classification. In few-shot settings, the model can learn the background from limited data. As the sample size increases, the improvements become more significant. However, due to the distinct fore-background differences in the dataset, the performance of the baseline model is already impressive, and improvements brought by BackMix are relatively limited. Note that BackMix only provides limited performance gains and will not affect the capability of the original method in this case.}

\revise{Practically, known and unknown classes are likely to appear in similar scenes or belong to classes without strong fore-background correlations. Therefore, BackMix is effective in real open scenarios.}

\revise{\textbf{Few-shot learning from scratch.} Since BackMix relies on class activation maps (CAMs) to segment and mask the foreground, limited model capacity can cause multiple foregrounds to appear in a single image. As we have verified in experiments (See Section~\ref{sec:pre_method}) when multiple foreground objects appear in the image and one-hot labels are used, the performance is weaker than those with only one foreground object.}
\begin{figure}[t]
\subfigure[]{  
        \includegraphics[width=0.225\textwidth]{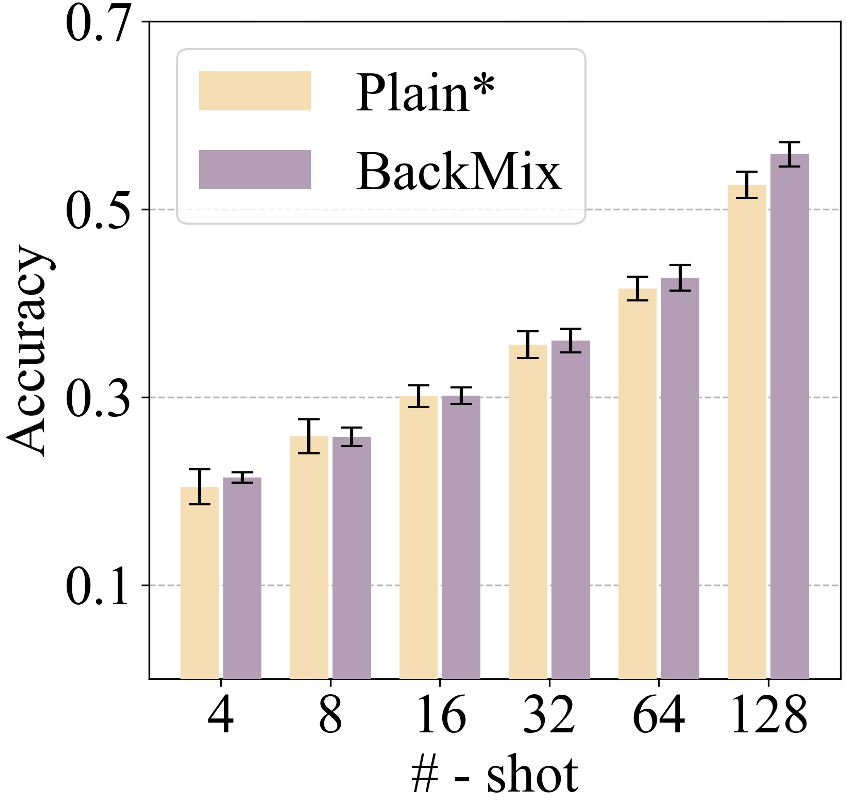}
        }
        \hfill
\subfigure[]{  
        \includegraphics[width=0.225\textwidth]{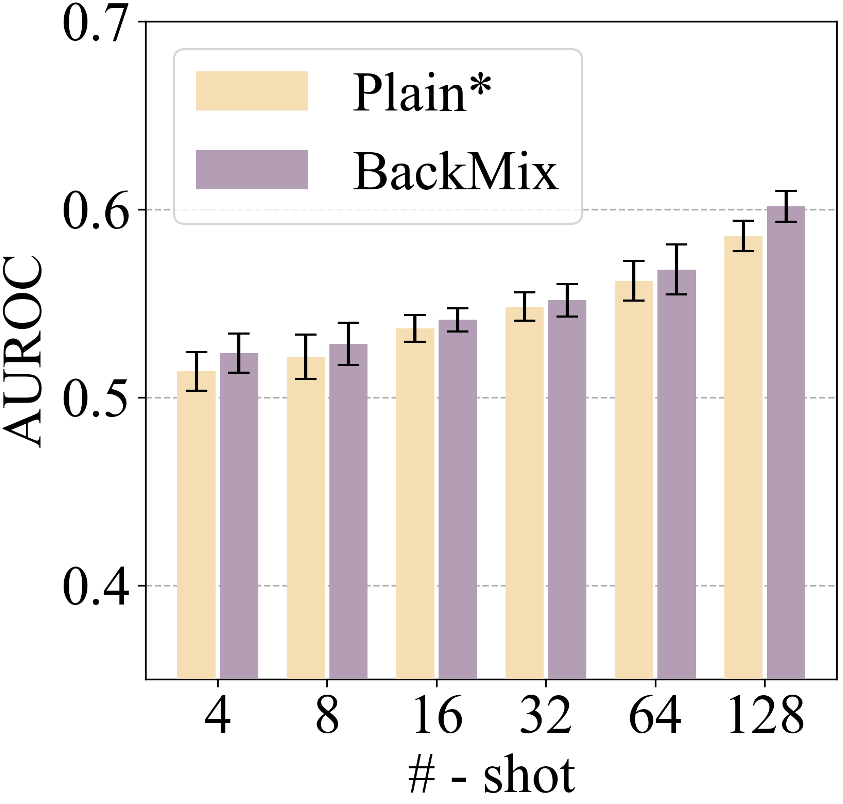}
        }
    \caption{\revise{Comparison of the (a) closed-set accuracy and (b) AUROC performance of the Plain* and the BackMix with varying numbers of the training samples from each known class. We conducted experiments under each setting with five different random seeds and reported the average results.}}
    \label{fig:fs_bar}
\end{figure}

\begin{figure*}[t]
\subfigure[]{  
        \includegraphics[width=0.98\textwidth]{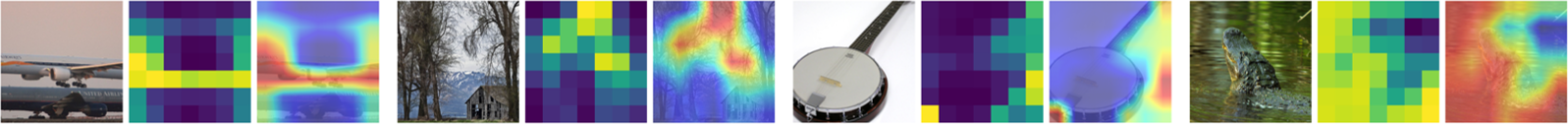}
        }
        \hfill
\subfigure[]{  
        \includegraphics[width=0.98\textwidth]{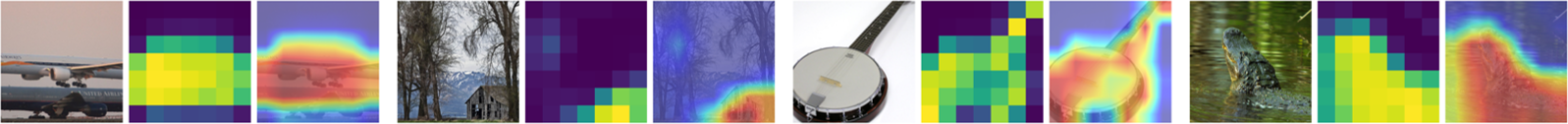}
        }
    \caption{\revise{Foreground segmentation masks learned by the model on the ImageNet30 dataset. Brighter areas indicate higher activation intensity, (a) under the 4-shot training setting, (b) under the 64-shot training setting.}}
    \label{fig:mask_fs}
\end{figure*}
\revise{Thus, in scenarios with few samples and no pretraining, it is difficult for the model to capture foreground regions and it may even use the image background for classification. We conducted the unknown detection experiment on the CIFAR10 dataset, varying the number of training samples per class from 4 to 128, to test the model's closed-set classification accuracy and unknown detection capability. Fig.~\ref{fig:fs_bar} shows that BackMix can only bring a slight or even no improvement when the number of training samples is minimal. As the number of samples increases, the gains of BackMix in both closed-set and open-set performance become more pronounced. In Fig.~\ref{fig:mask_fs}, we show the foreground segmentation of some images learned by the model when there are 4 training samples and 64 training samples per class, respectively. When the number of samples is too small, the learned segmentation of the foreground objects is inaccurate, resulting in multiple classified foregrounds in processed images and further confusing the model.}

\revise{This limitation can be addressed by finetuning a pretrained model that has a strong representation capacity with a small amount of downstream task data.}

\section{Details of the Datasets}
\label{sec:app_data_details}
\begin{itemize}
    \item \textbf{COCO~\cite{cocods}.} The COCO dataset includes 80 objects, 330,000 images, and 1.5 million object instances, which is widely used in target detection and segmentation. With its precise segmentation, we use it in Section~\ref{sec:3.1} for verifying unbalanced backgrounds may mislead the open set classifier. In Appendix~\ref{A_3_3}, we use it for testing the transferability of models trained with different data augmentation techniques on the image captioning task.
    \item \textbf{iNaturalist~\cite{Horn2018TheIS}.} iNaturalist is a natural image dataset from the real world. The sample distribution is different from the common data set used for classification, showing an extreme long-tailed distribution. Take advantage of its real-world features and use it in Section~\ref{sec:3.1} as an unknown class in the test phase.
    \item \textbf{CIFAR10~\cite{krizhevsky09learning}.} It includes 60,000 RGB images with the size of 32×32, among which 50,000 are divided into training set and the remaining 10,000 make up test set. This small-scale dataset includes ten common classes: airplane, automobile, bird, cat, deer, dog, frog, horse, ship, and truck, and each class has 6,000 images. It is used in Section~\ref{sec:3.2.2}, Section~\ref{sec:5.1}, Section~\ref{sec:5.3.1}, Section~\ref{sec:5.3.2}, Section~\ref{sec:5.3.3} and Appendix~\ref{A_3_1}, Appendix~\ref{A_3_2}, Appendix~\ref{A_3_4}, Appendix~\ref{A_4_2}.
    \item \textbf{LSUN-Crop/-Resize/-Fix.} The large-scale Scene Understanding dataset~\cite{yu2015lsun} includes images from ten different scenes, \eg~the kitchen, living room, and bedroom. It only has a test set of 10,000 images and can be reconstructed by cropping and resizing to LSUN-Crop and LSUN-Resize. LSUN-Fix is constructed by combining random sampling and resizing. They are regarded as unknown sets in Section~\ref{sec:3.2.2}, Section~\ref{sec:5.1.2}, Section~\ref{sec:5.3.2} and Appendix~\ref{A_3_2} and Appendix~\ref{A_3_5}.
    \item \textbf{ImageNet-Crop/-Resize/-Fix.} The ImageNet-Crop and ImageNet-Resize both have 10000 RGB images with the size of 32×32. Liang~\etal~\cite{liang2018enhancing} constructed them by cropping or downsampling the images from a subset of ImageNet~\cite{deng2009imagenet}. The ImageNet-Fix is from~\cite{tack2020csi}, which is constructed by randomly sampling and resizing the images in ImageNet. We use them in Section~\ref{sec:3.2.2}, Section~\ref{sec:5.1.2}, Section~\ref{sec:5.3.2} and Appendix~\ref{A_3_2}.
    \item \textbf{TinyImage~\cite{tinyimages80m}.} It consists of nearly 80 million tiny RGB images of size 32$\times$32 collected from the web. It is used as the outlier after removing its overlap with CIFAR10 in Section~\ref{sec:3.2.2}.
    \item \textbf{CIFAR100~\cite{krizhevsky09learning}.} CIFAR100 is a hierarchical dataset, which consists of RGB images size of 32$\times$32. Its 60,000 images are divided into 100 classes, each with 500 training images and 100 test images. We use it as the outlier in Section~\ref{sec:3.2.2} and see it as near OOD of CIFAR10 for OOD detection in Section~\ref{sec:5.1.3}, Section~\ref{sec:5.3.3} and Appendix~\ref{A_3_2}, Appendix~\ref{A_3_4}, Appendix~\ref{A_4_2}.
    \item \textbf{DTD~\cite{cimpoi14describing}.} The Describable Textures Dataset is a texture dataset consisting of 5,640 images across 47 classes, with each class containing 120 images. The dataset serves as a comprehensive resource for texture classification and recognition algorithms. We use it as an auxiliary outlier dataset that has limited semantic information in Section~\ref{sec:3.2.2}.
    \item \textbf{Flower102~\cite{Nilsback2008flower}.} The Oxford 102 Flower dataset comprises 102 classes of flowers. Each class contains between 40 and 258 images. We use it as an auxiliary outlier dataset that has limited semantic information in Section~\ref{sec:3.2.2}.
    \item \textbf{SVHN~\cite{netzer2011reading}.} The Street View House Number Dataset is derived from Google Street View house number and has images of digits 0-9. We tested the unknown detection capability of the model in this dataset in Section~\ref{sec:5.1.1}. It is used as the far OOD dataset of CIFAR10 in Section~\ref{sec:5.1.3} and used as the OOD dataset in Appendix~\ref{A_3_2}, Appendix~\ref{A_3_4}, and Appendix~\ref{A_3_5}. 
    \item \textbf{CIFAR+10.} The fixed openness of the model is also restricted due to the fixed data of known and unknown classes in Section~\ref{sec:5.1.1} experiments on CIFAR10. Following the protocol in~\cite{neal2018open}, we select four classes in CIFAR10 as known classes and select 10 classes from CIFAR100 as unknown classes. To avoid possible overlap of classes, we only select non-animal classes in CIFAR10 while only selecting from animal classes in CIFAR100.
    \item \textbf{CIFAR+50.} Like CIFAR+10, CIFAR+50 uses 50 animal classes from CIFAR100 as unknown classes. As the number of unknown classes increases, the openness correspondingly increases and becomes more challenging in Section~\ref{sec:5.1.1} experiments. We also use this setting in Appendix~\ref{A_3_1}.
    \item \textbf{Tiny-ImageNet~\cite{pouransari2014tiny}.} Tiny-ImageNet contains 100,000 RGB images of size 64×64. It is divided into 200 classes, each with 500 training images, 50 validation images, and 50 test images. Open set tasks also pose a greater challenge because of the richer classes available. In Section~\ref{sec:5.1.1} and Appendix~\ref{A_3_1}, we randomly select 20 classes as known classes and the remaining 180 classes as unknown classes.
    \item \textbf{ImageNet30~\cite{hendrycks2019using}.} ImageNet30 is composed of 30 low overlapped class images selected from ImageNet~\cite{deng2009imagenet}. Each class has 1300 training images and 100 test images. In Section~\ref{sec:5.2}, we take the first 10 classes as known classes and the last 20 as unknown classes in the dictionary order.
    \item  \textbf{ImageNet1K~\cite{deng2009imagenet}.}
    ImageNet1K dataset comprises 1000 classes with a total of approximately 1.2 million labeled images. It serves as a standard benchmark for image classification tasks and has been crucial in the development and evaluation of deep learning models. In Appendix~\ref{A_3_3}, we use all training samples for training the visual backbone.
    \item  \textbf{Pascal VOC~\cite{everingham2010pascal}.}
    The Visual Object Classes dataset is a significant benchmark in computer vision, primarily used for object detection and image classification. It includes a variety of object classes, such as animals, vehicles, and household items, and provides detailed annotations like bounding boxes and class labels. In Appendix~\ref{A_3_3}, we use this dataset for testing the performance of the model on the object detection task.
    \item \textbf{WaterBirds~\cite{sagawa2019distributionally}.}
    The Waterbirds dataset is a synthetic dataset that is commonly used to study the problem of spurious correlations. It uses bird images and corresponding segmentation annotations from the CUB~\cite{wah2011caltech} dataset, combined with background images from Places365~\cite{zhou2017places} dataset, to create two classes: water birds and land birds. The CUB dataset contains 200 classes of birds and the Places365 dataset is composed of 434 scene classes. It introduces a correlation between the type of bird and the background (water or land), which can mislead models to rely on the background for classification instead of focusing on the bird itself. In Appendix~\ref{A_3_5}, we use this dataset for testing the performance of the model against the spurious correlations in the training stage.
    \item \textbf{iSUN~\cite{xu2015turkergaze}.} The iSUN dataset contains a rich variety of natural scene images. In Appendix~\ref{A_3_5}, we use this dataset as the OOD dataset.
\end{itemize}
\bibliography{ref}
\bibliographystyle{IEEEtran}

\end{document}